\PassOptionsToPackage{round,authoryear}{natbib}
\PassOptionsToPackage{table,dvipsnames}{xcolor}
\documentclass[]{style/company_light} 

\usepackage[utf8]{inputenc}
\usepackage{hyperref}
\usepackage{url}
\usepackage{booktabs}
\usepackage{amsfonts}
\usepackage{amsmath}
\usepackage{nicefrac}
\usepackage{microtype}
\usepackage{graphicx}
\usepackage{subcaption}
\usepackage{multirow}
\usepackage{enumerate}
\usepackage{enumitem}
\usepackage[most]{tcolorbox}
\usepackage{adjustbox}
\usepackage{array}
\usepackage{amssymb}

\usepackage{booktabs}
\usepackage{xcolor}
\usepackage{colortbl}
\usepackage{adjustbox}

\usepackage{booktabs,longtable,array,multirow,makecell}
\usepackage[table]{xcolor}
\usepackage{booktabs,pifont,graphicx,xcolor}
\usepackage{booktabs,tabularx,array}

\definecolor{phyblue}{RGB}{144,190,220}

\NewDocumentCommand{\todo}
{ mO{} }{\textcolor{magenta}{\textsuperscript{\textit{TODO}}\textsf{\textbf{\small[#1]}}}}

\title{World Models’ Last Exam in Physics}
\author{Mingju Gao$^{*,\dagger}$, Qingle Liu$^{*}$, Yuzhao Peng$^{*}$, Xinjie Lin$^{*}$, Ziming Qin, Zheng Jiang, Wenyi Li, Calvin Xiao, Youjie Zheng, Kaisen Yang$^{\ddagger}$, Qinhuai Na$^{\ddagger}$}

\renewcommand\affiliationformat[2][]{\makebox[\linewidth][c]{\small\bfseries #2}}
\affiliation{Navers Lab, Einsia.AI\quad Peking University\quad Tsinghua University}
\renewcommand\contributionformat[2][]{%
  \vskip 0.15cm
  \makebox[\linewidth][c]{\footnotesize\color{gray}#2}%
}
\contribution{$^{*}$Equal Contribution\quad$^{\ddagger}$Corresponding Author\quad$^{\dagger}$Project Lead}

\abstract{%
Video world models can produce visually convincing yet physically inconsistent sequences, raising concerns about their reliability for prediction and planning in embodied AI systems.
Existing evaluations often rely on model-based judgments or reference videos, while direct physical tests largely focus on mechanics.
We introduce \emph{World Models' Last Exam in Physics}, a measurement-based benchmark for evaluating physical consistency in video world models.
The benchmark comprises 40 controlled tasks spanning mechanics, optics, fluids, thermal and phase-change phenomena, electromagnetism, and surface tension.
Each task pairs an initial image and a generation prompt with predefined physical criteria, enabling interpretable tests of observable physical relationships without requiring reference videos.
Its evaluator combines task-observability screening with task-specific quantitative physical measurements.
Experiments on eight video generation models across 1,280 videos reveal persistent physical inconsistencies and substantial variation across tasks, with the best model achieving an overall score of 57.76 out of 100.
Evaluation on synthetic videos with known physical relationships provides evidence for the validity of the measurement module under controlled conditions.
The evaluator also achieves higher agreement with human judgments than a direct vision-language model baseline in both within-task rankings and pairwise comparisons.
By combining coverage across physical domains with scores grounded in measurable evidence and explicit measurement limitations, the benchmark provides an interpretable basis for diagnosing physical inconsistencies and tracking progress toward physically consistent video world models.
}

\date{\today}
\metadata[HomePage]{\url{https://lab.einsia.ai/phys-last-exam}}
\metadata[Contact]{\href{mailto:yuno@einsia.ai}{\texttt{yuno@einsia.ai}}, \href{mailto:kaisenyang@einsia.ai}{\texttt{kaisenyang@einsia.ai}}, \href{mailto:nana@einsia.ai}{\texttt{nana@einsia.ai}}}

\begin{document}

\maketitle


\begin{figure*}[t]
    \centering
    \includegraphics[width=\linewidth]{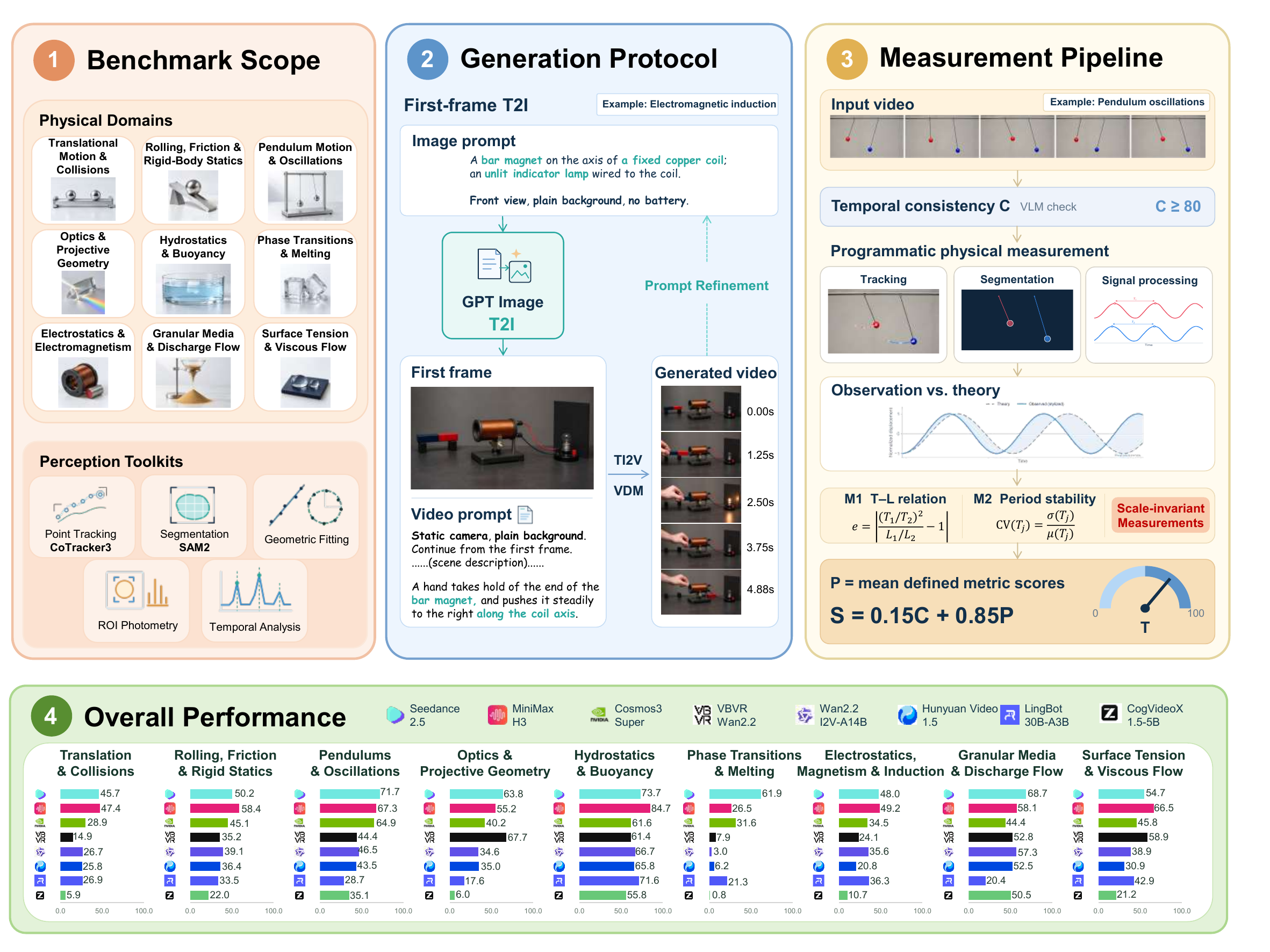}
    \caption{\textbf{Overview of our benchmark.}
    \textbf{(1) Benchmark scope:} 40 controlled tasks span nine physical categories, supported by perception tools for extracting observable quantities.
    \textbf{(2) Generation protocol:} Each task pairs a generated initial frame with a task-specific video prompt, refined during benchmark construction. The coil-and-magnet scene illustrates this process.
    \textbf{(3) Measurement pipeline:} A separate pendulum example illustrates consistency screening, physical measurement, and comparison with predefined physical criteria.
    \textbf{(4) Overall performance:} Mean composite scores summarize the performance of eight video generation models across the nine categories. Normalized composite scores are multiplied by 100 for presentation, with higher values indicating better performance.}
    \label{fig:teaser}
\end{figure*}

\section{Introduction}
\label{sec:introduction}

Using video generation models as world models for prediction and planning requires them to reproduce how environments evolve under physical constraints. Despite substantial advances in visual quality, temporal coherence, and scene complexity~\citep{bar2024lumiere,yang2025cogvideox,polyak2024moviegen,kong2024hunyuanvideo,wan2025wan,zheng2025opensora,ma2026lingbot,seedance2026seedance,team2025kling}, current models can still generate physically implausible motions and interactions~\citep{bansal2025videophy,meng2025phygenbench,radsch2026physics}. As video models are increasingly explored as learned world simulators~\citep{yang2023unisim,bruce2024genie,qian2026matrixgame,zhao2026phyworld} and as foundations for world action models (WAMs) that couple future-state prediction with action generation~\citep{bi2026motus,li2026causal,kim2026cosmos,ye2026world}, such violations may undermine the reliability of simulated outcomes and the plans derived from them. Assessing their suitability for this role therefore requires systematic evaluation of whether generated events satisfy the relevant physical relationships. This evaluation should identify which phenomena models reproduce reliably, characterize where they fail, and determine whether new methods reduce these failures, providing a basis for measuring progress toward physically consistent video world models.

Existing evaluations of physical consistency draw on vision-language model (VLM) judgments, reference-based comparisons, or direct tests of physical relationships. VLM-based methods cover diverse phenomena, but their judgments depend on the evaluator's own physical reasoning~\citep{bansal2025videophy,meng2025phygenbench,lin2026phyground,luo2026vwgbench}. Current VLMs still have limited capabilities in understanding physical relationships and reasoning about dynamic interactions, which can compromise the reliability of their evaluations. For example, PQSG reports that GPT-5.5 achieves 64.6\% accuracy on physics questions in FinePhyEval, compared with 88.4\% on object-level questions, illustrating the difficulty of judging physical behavior~\citep{pothiraj2026pqsg}. Reference-based approaches assess generated videos through comparisons with reference observations. Physics-IQ covers multiple physical domains, but evaluates similarity to reference videos rather than directly testing physical laws, making its scores sensitive to reference quality and video alignment~\citep{radsch2026physics}. Direct physical tests, including Principia, instead measure quantities such as object motion and assess whether they satisfy physical laws or the conditions specified in the prompt~\citep{thozhiyoor2025gravity,le2025newtonrewards,khanbayov2026physweep,thozhiyoor2026principia}. These tests mainly cover mechanics, while some measurement-based approaches also require real recordings, calibrated data, or simulator ground truth~\citep{wang2026gauge,jain2026rigidbench}. 
These gaps motivate a framework for testing observable physical relationships across domains, in which physical scores are derived from task-specific measurements without requiring reference videos or simulator ground truth, while a VLM provides preliminary consistency and observability screening.



To address this challenge, we introduce \emph{World Models' Last Exam in Physics}, a measurement-based benchmark comprising 40 controlled tasks spanning mechanics, optics, fluids, thermal and phase-change phenomena, electromagnetism, and surface-tension effects.
Our central idea is to pair controlled generation conditions with measurable physical criteria.
As shown in Figure~\ref{fig:teaser}, each task defines an image--prompt input, a target physical process, and observable relationships for evaluating the generated video.
Our evaluator assesses temporal consistency and task observability for each video. Independently of the screening outcome, it attempts to extract task-specific quantities, such as oscillation periods, reflection angles, and liquid levels, using tracking, segmentation, and temporal analysis. The recovered quantities are compared with predefined physical criteria. The screening outcome determines whether the physical score contributes to the composite score; passing the screen neither establishes physical correctness nor guarantees recovery of every required quantity.  The resulting composite scores summarize model performance across the benchmark’s nine task categories.

We evaluate eight video generation models across 40 tasks using 1,280 generated videos.
The best-performing model achieves an overall score of 57.76 out of 100, with substantial variation in performance across tasks and physical domains.
These results reveal persistent challenges in generating physically consistent videos and highlight the value of task-specific measurements for diagnosing model weaknesses.
Our contributions are threefold:
\begin{itemize}
    \item We introduce a measurement-based benchmark comprising 40 controlled video generation tasks covering mechanics, optics, fluid behavior, thermal and phase-change phenomena, electromagnetism, and surface-tension effects.
    \item We develop an interpretable evaluation protocol that combines temporal-consistency and task-observability screening with task-specific physical measurements, explicitly distinguishing physical test outcomes from cases with insufficient measurement evidence.
    \item We conduct a systematic evaluation of eight video generation models across the benchmark’s 40 tasks, revealing a gap between temporal coherence and performance on explicit physical tests, as well as substantial variation across tasks and physical domains.
\end{itemize}

\section{Method}
\label{sec:method}

\subsection{Overview and Task Formulation}
\label{sec:method_overview}

Our benchmark assesses observable physical consistency in generated videos through controlled experiments and quantitative measurements. It tests whether quantities extracted from a video, such as positions and angles, satisfy the physical relationships specified by the task. Each assessment is restricted to these observable relationships under the stated experimental assumptions; unmeasured physical properties remain unverified.

A task specifies physical assumptions $\mathcal{A}_k$, an image--prompt instance $(I_{0,k},p_k)$, and measurable criteria $\mathcal{R}_k$:
\begin{equation}
    \mathcal{T}_k=(\mathcal{A}_k,I_{0,k},p_k,\mathcal{R}_k),
    \qquad V_{k,r}=G_\theta(I_{0,k},p_k),
    \label{eq:method_pipeline}
\end{equation}
where $G_\theta$ is the evaluated video model and $V_{k,r}$ is the $r$-th video generated for task $k$. The assumptions describe materials, initial conditions, and observation geometry, while the criteria specify the observable physical relationships or events to evaluate. For example, a free-fall task assumes that an object is released from rest with negligible air resistance and checks whether its measured falling distance is proportional to the square of elapsed time.

\textbf{Overview.} As shown in Figure~\ref{fig:teaser}, the benchmark comprises 40 controlled tasks organized into nine task categories, evaluated through a standardized generation and measurement workflow. For each task, an initial frame is generated using GPT-Image-2.5 and paired with a prompt specifying how the experiment should proceed. The video model then generates a sequence conditioned on these inputs. Our evaluator first checks each generated video for temporal consistency and semantic coherence, assessing whether the required objects, events, and regions remain identifiable and measurable. Task-specific physical quantities are extracted using tools including CoTracker3~\citep{karaev2025cotracker3} for point tracking, SAM~2~\citep{ravi2025sam} for segmentation, and dedicated routines for geometric fitting, region-of-interest photometry, and temporal analysis. The extracted measurements are compared with the predefined physical criteria without requiring reference videos. The screening outcome determines whether the physical score contributes to the composite score, while independent measurement results and evidence availability are retained for diagnostic analysis. Composite scores are aggregated across tasks to compare model performance.

\subsection{Physics-Grounded Task Design}
\label{sec:method_tasks}

\begin{table*}[t]
\centering
\caption{The nine task categories, representative phenomena, and governing physical principles.}
\label{tab:task_principles}
\small
\setlength{\tabcolsep}{6pt}
\renewcommand{\arraystretch}{1.15}
\begin{tabularx}{\textwidth}{
    @{}
    >{\raggedright\arraybackslash}p{0.28\textwidth}
    >{\raggedright\arraybackslash}X
    >{\raggedright\arraybackslash}X
    @{}
}
\toprule
\textbf{Task category} & \textbf{Representative phenomena} & \textbf{Physical principles} \\
\midrule
Translational Motion and Collisions
& Free fall, projectiles, bouncing, and collisions
& Constant gravitational acceleration, ballistic motion, restitution, and momentum conservation \\
\addlinespace[3pt]
Rolling, Friction, and Rigid-Body Statics
& Rolling descent, sliding, tipping, and hanging chains
& Rolling constraints, rotational inertia, Coulomb friction, torque balance, and catenary equilibrium \\
\addlinespace[3pt]
Pendulum Motion and Oscillations
& Period dependence on amplitude, mass, and length
& Small-angle isochronism, mass independence, length scaling, and nonlinear pendulum dynamics \\
\addlinespace[3pt]
Optics and Projective Geometry
& Reflection, refraction, shadows, and marked-rod motion
& Reflection law, Snell's law, critical-angle condition, ray concurrency, and cross-ratio invariance \\
\addlinespace[3pt]
Hydrostatics and Buoyancy
& Liquid surfaces, communicating vessels, and floating ice
& Hydrostatic pressure balance, free-surface orientation, and Archimedes' principle \\
\addlinespace[3pt]
Phase Transitions and Melting
& Ice melting, freezing expansion, and comparative melting
& Mass conservation, buoyancy, density-dependent volume changes, and heat transfer \\
\addlinespace[3pt]
Electrostatics, Magnetism, and Electromagnetic Induction
& Charge repulsion, compass deflection, induction, and magnetic damping
& Electrostatic force balance, current-generated magnetic fields, Faraday's law, and Lenz's law \\
\addlinespace[3pt]
Granular Media and Discharge Flow
& Sandpile geometry and sand--water discharge comparisons
& Angle-of-repose consistency, Beverloo discharge scaling, and Torricelli's law \\
\addlinespace[3pt]
Surface Tension and Viscous Flow
& Capillary rise, connected bubbles, droplet merging, and viscous settling
& Jurin's law, Young--Laplace pressure, volume conservation, and Stokes drag \\
\bottomrule
\end{tabularx}
\end{table*}

\paragraph{Task coverage and design principle.}
We construct 40 tasks organized into nine categories, covering a broad range of common physical phenomena encountered in everyday life. Table~\ref{tab:task_principles} summarizes representative phenomena and their governing physical principles. Our central design principle is to formulate physical predictions as observable relationships that can be tested without absolute spatial or temporal calibration. We organize these tests into three types: spatial relationships among geometric quantities, temporal relationships among periods, durations, or event timings, and coupled spatiotemporal relationships between motion and geometry. For each task, we identify the governing law and its assumptions, derive the relationship to be tested, and design a scene that makes the required observables measurable within a single video. Where appropriate, matched comparative setups allow shared unknown factors to cancel from the tested relationship.

\textbf{Spatial relationships.}
Spatial tests check whether geometric quantities extracted from a physical process satisfy its predicted relationships. A classical example is projectile motion. Under uniform gravity and negligible air resistance, a projectile launched at $45^\circ$ and returning to its launch height should have a maximum height $H$ and horizontal range $R$ satisfying $H/R = 1/4$. Here, $H$ is measured relative to the launch height, and $R$ is the horizontal distance between launch and landing. In the oblique-projectile task, we evaluate
\begin{equation}
    r_{\mathrm{spatial}}
    =
    \left| \frac{H}{R} - \frac{1}{4} \right|.
\end{equation}
A residual near zero indicates agreement with the predicted trajectory geometry. The initial speed and gravitational acceleration cancel from this ratio, allowing the relationship to be tested without estimating either quantity. A common spatial scale also allows $H$ and $R$ to be measured directly in pixels without absolute length calibration. However, the ratio does not remove arbitrary perspective distortion. We therefore specify an appropriate side-view geometry in both the image-generation and video-generation prompts.


\paragraph{Temporal relationships.}
Temporal tests examine whether measured periods, durations, or event timings satisfy the expected physical relationship. For example, two small-angle pendulums with equal lengths under the same gravity should have approximately equal periods, even when their bob masses differ. We measure their periods $T_1$ and $T_2$ in frames and evaluate
\begin{equation}
    r_{\mathrm{temporal}}
    =\left|\frac{T_1}{T_2}-1\right|.
    \label{eq:method_temporal}
\end{equation}
This test checks whether changing the bob mass preserves the oscillation period. Because both periods are measured on a shared, uniformly sampled timeline, the common conversion from frames to time cancels in their ratio. We specify uniform temporal progression in the video-generation prompt to support the shared temporal-scale assumption.

\paragraph{Coupled spatiotemporal relationships.}
Spatiotemporal tests examine whether motion measurements are consistent with object geometry. For example, pure rolling without slipping requires the translational speed $v$ to equal the angular speed $\omega$ multiplied by the radius $R$. We measure $v$ in pixels per frame, $\omega$ in radians per frame, and $R$ in pixels, and evaluate
\begin{equation}
    r_{\mathrm{spatiotemporal}}
    =\left|\frac{v}{\omega R}-1\right|.
    \label{eq:method_spatiotemporal}
\end{equation}
This test checks whether the observed translation matches the rotation of a visible surface marker. Because $v$ and $\omega R$ share the same spatial and temporal scale factors, both unknown conversions cancel.

\subsection{First-Frame Construction and Video Generation}
\label{sec:method_inputs}

Reliable quantitative evaluation requires clearly defined experimental conditions and observable quantities for measurement. Ambiguous initial geometry, occlusion, or an incomplete observation interval can make a physical test inconclusive and confound model errors with limitations of the input setup. We therefore construct task-specific first frames and video prompts through an iterative process, as illustrated in Figure~\ref{fig:teaser}, to establish the prescribed initial configuration and make the target process observable.

For each task, we translate the experimental requirements into an image prompt and use GPT-Image-2.5 to generate candidate first frames that establish the initial configuration and expose the quantities needed for measurement. Each first frame is paired with a video prompt specifying how the experiment begins and unfolds, which scene properties remain fixed, and the observation interval required for evaluation. For example, pendulum experiments require fixed pivots and sufficient oscillations to estimate periods, while melting experiments require visible initial and final liquid levels. We iteratively refine both prompts through pilot generation and human inspection: ambiguous configurations or poorly visible objects lead to image-prompt revisions and regenerated first frames, while unclear actions or insufficient observation intervals lead to video-prompt revisions. Revised pairs are tested again, forming a feedback loop between input construction and pilot inspection. Errors in extracting clearly visible evidence are addressed separately through evaluator review. The finalized image--prompt pairs are then submitted to eight video models through a common generation interface. The evaluation covers 40 tasks across nine categories, yielding 1,280 videos with four samples per task--model pair.

\subsection{Physical Evaluation Protocol}
\label{sec:method_evaluation}

Generated videos do not always provide the evidence required for a meaningful physical test. For example, estimating free-fall acceleration requires a consistently identifiable ball and a recoverable trajectory. If the ball disappears, changes identity, or moves discontinuously in a way that prevents reliable tracking, the required measurements become unavailable. We therefore assess temporal consistency as a screening signal for score aggregation. As illustrated in Figure~\ref{fig:teaser}, we use Qwen3.6-27B~\citep{yang2025qwen3} to assign a temporal consistency score $C(V) \in [0,100]$, considering object persistence and visual continuity. Physical measurements are attempted independently for all available videos, regardless of $C(V)$. For videos with $C(V) < 80$, the physical score $P$ does not contribute to the composite score $S$, while independently obtained measurements and physical scores are retained. Passing the screen neither establishes physical correctness nor guarantees recovery of every required quantity.


Temporal consistency alone does not establish physical correctness: a clearly visible, continuously tracked object can still follow an incorrect trajectory or violate a motion--geometry relationship. We therefore compare recovered observables with task-specific physical predictions. For each task $k$, an extractor $E_k$ uses segmentation, tracking, geometric fitting, and event detection to recover quantities such as trajectories, periods, ray angles, contours, and liquid levels over the prescribed observation interval. For each indicator, we compute
\begin{equation}
    \mathbf z_k=E_k(V),
    \qquad
    r_{k,j}=f_{k,j}(\mathbf z_k;\mathcal A_k),
    \qquad
    q_{k,j}=\phi_{k,j}(r_{k,j})\in[0,1],
    \label{eq:method_measurement}
\end{equation}
where $\mathcal A_k$ denotes the experimental assumptions, $f_{k,j}$ computes a physical residual or event statistic, and $\phi_{k,j}$ maps it to an agreement score. Larger residuals receive lower agreement, while directional and event-based indicators follow their corresponding definitions. The primary indicator evaluates the target relationship, and auxiliary indicators provide complementary checks. After task-specific handling of unavailable measurements, the indicators are combined into a physics score:
\begin{equation}
    P_k(V)=100\,F_k(\mathbf q_k),
    \qquad
    \mathbf q_k=(q_{k,j})_{j\in\mathcal J_k},
    \label{eq:method_physics}
\end{equation}
where $\mathcal J_k$ contains the task's indicators and $F_k$ is its aggregation rule. Intermediate measurements and visual evidence are retained to help distinguish extraction errors from physical violations.

To summarize overall performance, we combine the consistency score with a gated physical-score contribution for every available video. Physical measurements are attempted independently of the gate, but the physical score contributes to the composite score only when $C(V) \geq 80$ and a physics score is recorded:
\begin{equation}
    S_k(V)=0.15C(V)+0.85\widetilde P_k(V),
    \qquad
    \widetilde P_k(V)=
    \begin{cases}
        P_k(V), & C(V)\geq80\ \text{and a physics score is recorded},\\
        0, & \text{otherwise}.
    \end{cases}
    \label{eq:method_total}
\end{equation}
For brevity, we write $S$ for $S_k(V)$ when the task and video are clear from context. Physical agreement receives the larger weight, while videos failing the screening receive only the temporal-consistency contribution. For each model, temporal consistency and total scores are averaged over all available videos. We average all recorded physics scores, including zeros, for each model and each category or difficulty level.

\section{Experiments}
\subsection{Settings}


\textbf{Video Generation Models.} We evaluate eight image-to-video models:\footnote{For each model, the values in parentheses specify the video resolution (width $\times$ height, in pixels) and the number of frames.} CogVideoX1.5-5B ($832 \times 480$, 81 frames), Cosmos3-Super ($832 \times 480$, 81 frames), HunyuanVideo-1.5 ($1264 \times 720$, 129 frames), LingBot-Video ($832 \times 480$, 81 frames), MiniMax-H3 ($1344 \times 768$, 124 frames), Seedance-2.5 ($1270 \times 726$, 121 frames), VBVR-Wan2.2 ($832 \times 480$, 81 frames), and Wan2.2-I2V-A14B ($832 \times 464$, 81 frames)~\citep{yang2025cogvideox, agarwal2026cosmos, wu2025hunyuanvideo, ma2026lingbot, wang2026very, wan2025wan, minimax2026h3, seed2026seedance25}. For each of the 40 tasks, all models receive the same initial frame and task description. We generate four samples per model--task pair, giving a planned total of 1,280 videos. Random seeds are fixed to 42--45 where supported.
 
\textbf{Metrics.} Each video is evaluated for task consistency and physical correctness. The consistency evaluator uses a dedicated prompt for each task to assess whether the required subjects, events, and comparisons are observable. Task-specific physical evaluators extract trajectories, periods, angles, and event timings using color segmentation, connected-component analysis, optical flow, and contour or line fitting, with SAM2~\citep{ravi2025sam} and CoTracker~\citep{karaev2025cotracker3} providing object segmentation and point tracking where needed. The extracted measurements are compared with the corresponding physical constraints to compute task-specific errors and scores. Most residual-based metrics map an error $e$ to a score through
\begin{equation}
s(e;a)=\frac{1}{1+|e|/a},
\end{equation}
where $a>0$ is a predefined, metric-specific error scale. For example, the pendulum-length task uses $a=0.20$ and computes
\begin{equation}
e=\left|\frac{(T_1/T_2)^2}{L_1/L_2}-1\right|,
\end{equation}
where $T_1,T_2$ and $L_1,L_2$ are the periods and lengths of the two pendulums, respectively. Other indicators use task-specific mappings, including binary directional tests and event-evidence scores. The physical score is the arithmetic mean of the indicators defined for each task. All normalization parameters remain fixed across generative models.

\subsection{Main Results}

\begin{table*}[tp]
\centering
\newsavebox{\VDMTableBox}
\begin{lrbox}{\VDMTableBox}
\begin{minipage}{\textwidth}
\centering
\normalfont\rmfamily

\definecolor{VDMMuted}{HTML}{666666}
\definecolor{VDMPassGreen}{HTML}{287A43}
\definecolor{VDMFailRed}{HTML}{BE3D3D}
\definecolor{VDMLow}{HTML}{F4DCDC}
\definecolor{VDMMid}{HTML}{FAF0C9}
\definecolor{VDMHigh}{HTML}{DEEEDA}
\providecommand{\VDMLogoPath}{Figures/vdm_logos}

\newcommand{\VDMCheck}{\textcolor{VDMPassGreen}{\ding{51}}}
\newcommand{\VDMCross}{\textcolor{VDMFailRed}{\ding{55}}}
\newcommand{\VDMStatus}[2]{%
  \ifnum#1=#2\relax\VDMCheck\else\VDMCross\fi}
\newcommand{\VDMCell}[4]{%
  \cellcolor{#4}%
  \makebox[3.35em][r]{#1}\kern.10em%
  \makebox[.85em][c]{\VDMStatus{#2}{#3}}}
\newcommand{\VDMAvgCell}[2]{%
  \cellcolor{#2}%
  \makebox[3.35em][r]{#1}\kern.10em\makebox[.85em]{}}
\newcommand{\VDMTask}[2]{%
  \mbox{\makebox[2.8em][l]{\textcolor{VDMMuted}{#1}}#2}}
\newcommand{\VDMGroup}[1]{%
  \addlinespace[2.4pt]
  \multicolumn{10}{@{}l}{%
    \normalfont\rmfamily\bfseries\fontsize{7.1}{8.2}\selectfont #1}\\[-.6pt]}
\newcommand{\VDMLogo}[1]{%
  \parbox[c][6.2mm][c]{9mm}{\centering
    \IfFileExists{\VDMLogoPath/#1.png}{%
      \includegraphics[width=7mm,height=4.8mm,keepaspectratio]{\VDMLogoPath/#1.png}%
    }{\mbox{}}}}
\newsavebox{\VDMModelBox}
\newcommand{\VDMModel}[1]{%
  \sbox{\VDMModelBox}{\normalfont\rmfamily\bfseries\fontsize{7.1}{8.2}\selectfont #1}%
  \ifdim\wd\VDMModelBox>\linewidth
    \resizebox{\linewidth}{!}{\usebox{\VDMModelBox}}%
  \else\usebox{\VDMModelBox}\fi}
\newcommand{\VDMVersion}[1]{%
  \textcolor{VDMMuted}{\normalfont\rmfamily\fontsize{6.4}{7.2}\selectfont #1}}

\setlength{\abovecaptionskip}{0pt}
\setlength{\belowcaptionskip}{4pt}
\caption[Main results under the automatic consistency gate]{%
\textbf{Main results under the automatic consistency gate.}
Composite scores (0--100; higher is better) for 40 tasks and 1,280 available videos.
Bold marks row maxima, including ties at the displayed precision; backgrounds run from red (0) through yellow (50) to green (100).
Within each \emph{model--task pair}, \VDMCheck{} means every available video has $C\ge80$; \VDMCross{} means at least one has $C<80$.
Per-video scores use $S=0.15C+0.85P\,\mathbf{1}[C\ge80]$, where $C$ is the original automatic consistency score and $P$ the independently recorded physical score.
E/M/H are empirical difficulty groups based on cross-model task-mean scores (15/15/10 tasks).}
\label{tab:vdm-task-model}

\begingroup
\normalfont\rmfamily\fontsize{7.5}{8.5}\selectfont
\setlength{\tabcolsep}{1.8pt}
\renewcommand{\arraystretch}{1.02}
\setlength{\heavyrulewidth}{.75pt}
\setlength{\lightrulewidth}{.35pt}
\newlength{\VDMContentWidth}
\setlength{\VDMContentWidth}{\dimexpr\textwidth-18\tabcolsep\relax}
\begin{tabular}{@{}
  >{\raggedright\arraybackslash}p{.315\VDMContentWidth}
  >{\centering\arraybackslash}p{.04\VDMContentWidth}
  *{8}{>{\centering\arraybackslash}p{.080625\VDMContentWidth}}
@{}}
\toprule
 & &
 \VDMLogo{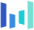} & \VDMLogo{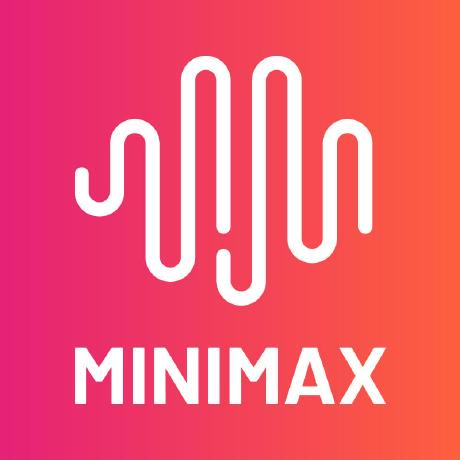} & \VDMLogo{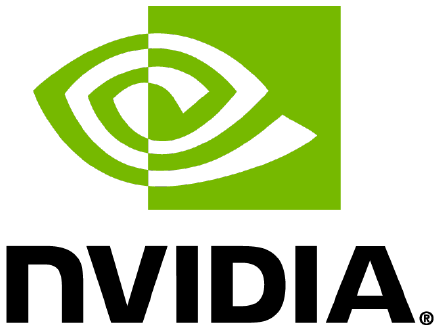} &
 \VDMLogo{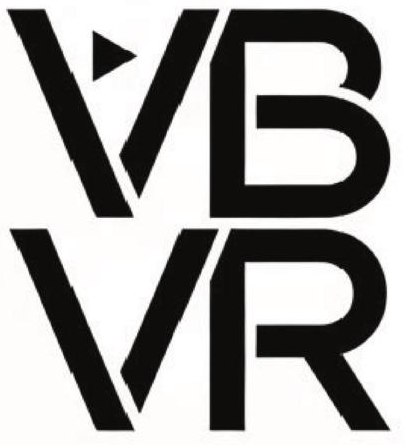} & \VDMLogo{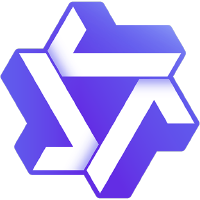} & \VDMLogo{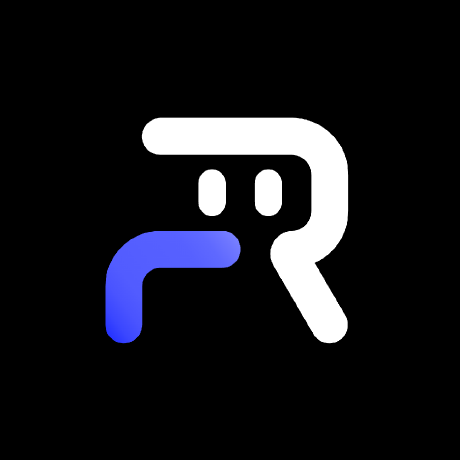} &
 \VDMLogo{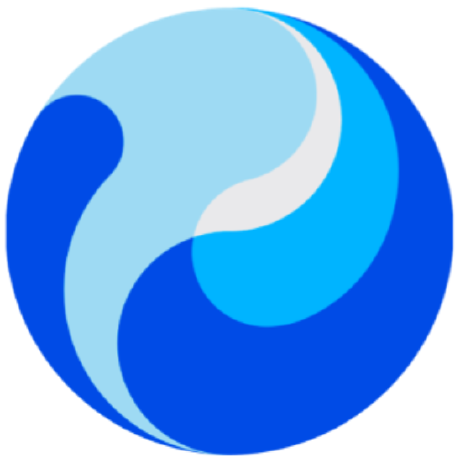} & \VDMLogo{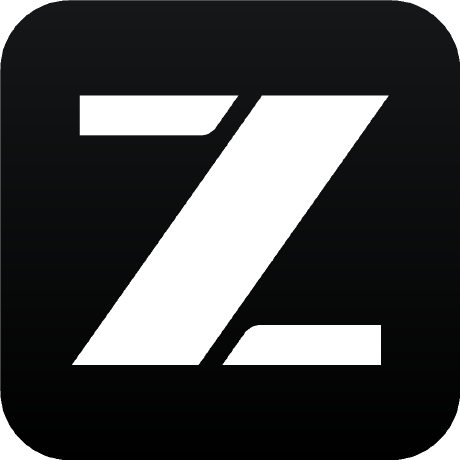}\\[-.7pt]
\multicolumn{1}{@{}c}{\VDMModel{Task}} & \VDMVersion{Level} &
 \VDMModel{Seedance} & \VDMModel{MiniMax} & \VDMModel{Cosmos} &
 \VDMModel{VBVR} & \VDMModel{Wan} & \VDMModel{LingBot} &
 \VDMModel{Hunyuan} & \VDMModel{CogVideoX}\\[-.7pt]
 & &
 \VDMVersion{2.5} & \VDMVersion{H3} & \VDMVersion{3 Super} &
 \VDMVersion{Wan2.2} & \VDMVersion{2.2-A14B} & \VDMVersion{30B-A3B} &
 \VDMVersion{1.5} & \VDMVersion{1.5-5B}\\[1.3pt]
\midrule

\VDMGroup{1. Translational Motion and Collisions}

\VDMTask{P1}{Bounce-height decay} & E &
 \VDMCell{\textbf{90.76}}{4}{4}{VDMHigh!81.52!VDMMid} &
 \VDMCell{88.19}{4}{4}{VDMHigh!76.38!VDMMid} &
 \VDMCell{57.46}{4}{4}{VDMHigh!14.92!VDMMid} &
 \VDMCell{0.00}{0}{4}{VDMMid!0.00!VDMLow} &
 \VDMCell{85.66}{4}{4}{VDMHigh!71.32!VDMMid} &
 \VDMCell{48.67}{4}{4}{VDMMid!97.34!VDMLow} &
 \VDMCell{83.03}{4}{4}{VDMHigh!66.06!VDMMid} &
 \VDMCell{3.75}{1}{4}{VDMMid!7.50!VDMLow} \\

\VDMTask{P2}{Free fall} & M &
 \VDMCell{29.50}{4}{4}{VDMMid!59.00!VDMLow} &
 \VDMCell{25.84}{4}{4}{VDMMid!51.68!VDMLow} &
 \VDMCell{\textbf{34.54}}{4}{4}{VDMMid!69.08!VDMLow} &
 \VDMCell{24.27}{4}{4}{VDMMid!48.54!VDMLow} &
 \VDMCell{22.72}{4}{4}{VDMMid!45.44!VDMLow} &
 \VDMCell{34.41}{4}{4}{VDMMid!68.82!VDMLow} &
 \VDMCell{23.60}{4}{4}{VDMMid!47.20!VDMLow} &
 \VDMCell{18.03}{4}{4}{VDMMid!36.06!VDMLow} \\

\VDMTask{P3}{Complementary-angle throws} & M &
 \VDMCell{53.96}{4}{4}{VDMHigh!7.92!VDMMid} &
 \VDMCell{\textbf{57.66}}{4}{4}{VDMHigh!15.32!VDMMid} &
 \VDMCell{20.97}{4}{4}{VDMMid!41.94!VDMLow} &
 \VDMCell{10.88}{3}{4}{VDMMid!21.76!VDMLow} &
 \VDMCell{0.00}{0}{4}{VDMMid!0.00!VDMLow} &
 \VDMCell{30.87}{2}{4}{VDMMid!61.74!VDMLow} &
 \VDMCell{3.38}{1}{4}{VDMMid!6.76!VDMLow} &
 \VDMCell{0.00}{0}{4}{VDMMid!0.00!VDMLow} \\

\VDMTask{P4}{Projectile motion} & H &
 \VDMCell{29.13}{4}{4}{VDMMid!58.26!VDMLow} &
 \VDMCell{\textbf{39.01}}{4}{4}{VDMMid!78.02!VDMLow} &
 \VDMCell{17.25}{3}{4}{VDMMid!34.50!VDMLow} &
 \VDMCell{22.33}{3}{4}{VDMMid!44.66!VDMLow} &
 \VDMCell{10.36}{1}{4}{VDMMid!20.72!VDMLow} &
 \VDMCell{15.43}{3}{4}{VDMMid!30.86!VDMLow} &
 \VDMCell{7.83}{1}{4}{VDMMid!15.66!VDMLow} &
 \VDMCell{0.00}{0}{4}{VDMMid!0.00!VDMLow} \\

\VDMTask{P5}{Equal-mass collision} & H &
 \VDMCell{25.37}{4}{4}{VDMMid!50.74!VDMLow} &
 \VDMCell{\textbf{26.06}}{4}{4}{VDMMid!52.12!VDMLow} &
 \VDMCell{14.50}{3}{3}{VDMMid!29.00!VDMLow} &
 \VDMCell{17.05}{3}{4}{VDMMid!34.10!VDMLow} &
 \VDMCell{14.79}{3}{4}{VDMMid!29.58!VDMLow} &
 \VDMCell{4.97}{1}{4}{VDMMid!9.94!VDMLow} &
 \VDMCell{11.25}{3}{4}{VDMMid!22.50!VDMLow} &
 \VDMCell{7.50}{2}{4}{VDMMid!15.00!VDMLow} \\

\VDMGroup{2. Rolling, Friction, and Rigid-Body Statics}

\VDMTask{P6}{Mass-independent sliding} & E &
 \VDMCell{66.00}{4}{4}{VDMHigh!32.00!VDMMid} &
 \VDMCell{97.89}{4}{4}{VDMHigh!95.78!VDMMid} &
 \VDMCell{\textbf{98.46}}{4}{4}{VDMHigh!96.92!VDMMid} &
 \VDMCell{51.12}{4}{4}{VDMHigh!2.24!VDMMid} &
 \VDMCell{50.98}{4}{4}{VDMHigh!1.96!VDMMid} &
 \VDMCell{57.50}{4}{4}{VDMHigh!15.00!VDMMid} &
 \VDMCell{36.06}{4}{4}{VDMMid!72.12!VDMLow} &
 \VDMCell{43.62}{2}{4}{VDMMid!87.24!VDMLow} \\

\VDMTask{P7}{Hanging-chain equilibrium} & E &
 \VDMCell{68.85}{4}{4}{VDMHigh!37.70!VDMMid} &
 \VDMCell{70.29}{4}{4}{VDMHigh!40.58!VDMMid} &
 \VDMCell{71.09}{4}{4}{VDMHigh!42.18!VDMMid} &
 \VDMCell{69.71}{4}{4}{VDMHigh!39.42!VDMMid} &
 \VDMCell{71.26}{4}{4}{VDMHigh!42.52!VDMMid} &
 \VDMCell{69.11}{4}{4}{VDMHigh!38.22!VDMMid} &
 \VDMCell{\textbf{76.95}}{4}{4}{VDMHigh!53.90!VDMMid} &
 \VDMCell{69.60}{4}{4}{VDMHigh!39.20!VDMMid} \\

\VDMTask{P8}{Solid-sphere rolling} & M &
 \VDMCell{\textbf{75.56}}{4}{4}{VDMHigh!51.12!VDMMid} &
 \VDMCell{45.82}{4}{4}{VDMMid!91.64!VDMLow} &
 \VDMCell{28.79}{4}{4}{VDMMid!57.58!VDMLow} &
 \VDMCell{31.53}{4}{4}{VDMMid!63.06!VDMLow} &
 \VDMCell{35.39}{4}{4}{VDMMid!70.78!VDMLow} &
 \VDMCell{18.77}{4}{4}{VDMMid!37.54!VDMLow} &
 \VDMCell{37.11}{4}{4}{VDMMid!74.22!VDMLow} &
 \VDMCell{14.78}{3}{4}{VDMMid!29.56!VDMLow} \\

\VDMTask{P9}{Solid sphere vs. hoop} & M &
 \VDMCell{32.01}{4}{4}{VDMMid!64.02!VDMLow} &
 \VDMCell{40.83}{4}{4}{VDMMid!81.66!VDMLow} &
 \VDMCell{30.36}{4}{4}{VDMMid!60.72!VDMLow} &
 \VDMCell{\textbf{41.63}}{4}{4}{VDMMid!83.26!VDMLow} &
 \VDMCell{17.28}{4}{4}{VDMMid!34.56!VDMLow} &
 \VDMCell{22.08}{4}{4}{VDMMid!44.16!VDMLow} &
 \VDMCell{39.36}{4}{4}{VDMMid!78.72!VDMLow} &
 \VDMCell{3.75}{1}{4}{VDMMid!7.50!VDMLow} \\

\VDMTask{P10}{Edge-pivot toppling} & M &
 \VDMCell{34.27}{4}{4}{VDMMid!68.54!VDMLow} &
 \VDMCell{\textbf{66.53}}{4}{4}{VDMHigh!33.06!VDMMid} &
 \VDMCell{34.03}{4}{4}{VDMMid!68.06!VDMLow} &
 \VDMCell{4.76}{1}{4}{VDMMid!9.52!VDMLow} &
 \VDMCell{43.23}{4}{4}{VDMMid!86.46!VDMLow} &
 \VDMCell{14.38}{3}{4}{VDMMid!28.76!VDMLow} &
 \VDMCell{18.99}{4}{4}{VDMMid!37.98!VDMLow} &
 \VDMCell{0.00}{0}{4}{VDMMid!0.00!VDMLow} \\

\VDMTask{P11}{Rough-incline round trip} & H &
 \VDMCell{24.43}{4}{4}{VDMMid!48.86!VDMLow} &
 \VDMCell{\textbf{28.84}}{4}{4}{VDMMid!57.68!VDMLow} &
 \VDMCell{8.10}{2}{4}{VDMMid!16.20!VDMLow} &
 \VDMCell{12.36}{2}{4}{VDMMid!24.72!VDMLow} &
 \VDMCell{16.53}{2}{4}{VDMMid!33.06!VDMLow} &
 \VDMCell{19.18}{2}{4}{VDMMid!38.36!VDMLow} &
 \VDMCell{10.10}{2}{4}{VDMMid!20.20!VDMLow} &
 \VDMCell{0.00}{0}{4}{VDMMid!0.00!VDMLow} \\

\VDMGroup{3. Pendulum Motion and Oscillations}

\VDMTask{P12}{Pendulum period vs. mass} & E &
 \VDMCell{\textbf{80.41}}{4}{4}{VDMHigh!60.82!VDMMid} &
 \VDMCell{72.58}{4}{4}{VDMHigh!45.16!VDMMid} &
 \VDMCell{68.16}{4}{4}{VDMHigh!36.32!VDMMid} &
 \VDMCell{56.28}{3}{4}{VDMHigh!12.56!VDMMid} &
 \VDMCell{35.56}{4}{4}{VDMMid!71.12!VDMLow} &
 \VDMCell{19.25}{3}{4}{VDMMid!38.50!VDMLow} &
 \VDMCell{52.08}{4}{4}{VDMHigh!4.16!VDMMid} &
 \VDMCell{40.38}{4}{4}{VDMMid!80.76!VDMLow} \\

\VDMTask{P13}{Large-angle pendulum} & E &
 \VDMCell{57.76}{4}{4}{VDMHigh!15.52!VDMMid} &
 \VDMCell{\textbf{79.88}}{4}{4}{VDMHigh!59.76!VDMMid} &
 \VDMCell{62.81}{4}{4}{VDMHigh!25.62!VDMMid} &
 \VDMCell{43.59}{3}{4}{VDMMid!87.18!VDMLow} &
 \VDMCell{53.77}{4}{4}{VDMHigh!7.54!VDMMid} &
 \VDMCell{46.92}{4}{4}{VDMMid!93.84!VDMLow} &
 \VDMCell{73.52}{4}{4}{VDMHigh!47.04!VDMMid} &
 \VDMCell{48.41}{3}{4}{VDMMid!96.82!VDMLow} \\

\VDMTask{P14}{Pendulum period vs. length} & E &
 \VDMCell{66.15}{4}{4}{VDMHigh!32.30!VDMMid} &
 \VDMCell{65.41}{4}{4}{VDMHigh!30.82!VDMMid} &
 \VDMCell{61.93}{4}{4}{VDMHigh!23.86!VDMMid} &
 \VDMCell{55.83}{4}{4}{VDMHigh!11.66!VDMMid} &
 \VDMCell{\textbf{67.11}}{4}{4}{VDMHigh!34.22!VDMMid} &
 \VDMCell{25.36}{4}{4}{VDMMid!50.72!VDMLow} &
 \VDMCell{16.00}{2}{4}{VDMMid!32.00!VDMLow} &
 \VDMCell{26.03}{3}{4}{VDMMid!52.06!VDMLow} \\

\VDMTask{P15}{Small-angle isochronism} & M &
 \VDMCell{\textbf{82.60}}{4}{4}{VDMHigh!65.20!VDMMid} &
 \VDMCell{51.35}{4}{4}{VDMHigh!2.70!VDMMid} &
 \VDMCell{66.77}{4}{4}{VDMHigh!33.54!VDMMid} &
 \VDMCell{21.90}{1}{4}{VDMMid!43.80!VDMLow} &
 \VDMCell{29.69}{4}{4}{VDMMid!59.38!VDMLow} &
 \VDMCell{23.40}{3}{4}{VDMMid!46.80!VDMLow} &
 \VDMCell{32.52}{3}{4}{VDMMid!65.04!VDMLow} &
 \VDMCell{25.64}{4}{4}{VDMMid!51.28!VDMLow} \\

\VDMGroup{4. Optics and Projective Geometry}

\VDMTask{P16}{Collinear-point cross-ratio} & E &
 \VDMCell{81.10}{4}{4}{VDMHigh!62.20!VDMMid} &
 \VDMCell{82.97}{4}{4}{VDMHigh!65.94!VDMMid} &
 \VDMCell{82.39}{4}{4}{VDMHigh!64.78!VDMMid} &
 \VDMCell{83.92}{4}{4}{VDMHigh!67.84!VDMMid} &
 \VDMCell{68.12}{4}{4}{VDMHigh!36.24!VDMMid} &
 \VDMCell{43.56}{2}{4}{VDMMid!87.12!VDMLow} &
 \VDMCell{\textbf{88.34}}{4}{4}{VDMHigh!76.68!VDMMid} &
 \VDMCell{3.75}{1}{4}{VDMMid!7.50!VDMLow} \\

\VDMTask{P17}{Light refraction} & M &
 \VDMCell{65.59}{4}{4}{VDMHigh!31.18!VDMMid} &
 \VDMCell{65.68}{4}{4}{VDMHigh!31.36!VDMMid} &
 \VDMCell{56.38}{4}{4}{VDMHigh!12.76!VDMMid} &
 \VDMCell{\textbf{66.18}}{4}{4}{VDMHigh!32.36!VDMMid} &
 \VDMCell{57.94}{4}{4}{VDMHigh!15.88!VDMMid} &
 \VDMCell{14.62}{4}{4}{VDMMid!29.24!VDMLow} &
 \VDMCell{15.00}{4}{4}{VDMMid!30.00!VDMLow} &
 \VDMCell{15.00}{4}{4}{VDMMid!30.00!VDMLow} \\

\VDMTask{P18}{Light reflection} & M &
 \VDMCell{\textbf{96.15}}{4}{4}{VDMHigh!92.30!VDMMid} &
 \VDMCell{49.82}{3}{4}{VDMMid!99.64!VDMLow} &
 \VDMCell{0.00}{0}{3}{VDMMid!0.00!VDMLow} &
 \VDMCell{95.75}{4}{4}{VDMHigh!91.50!VDMMid} &
 \VDMCell{0.00}{0}{4}{VDMMid!0.00!VDMLow} &
 \VDMCell{3.38}{1}{4}{VDMMid!6.76!VDMLow} &
 \VDMCell{24.73}{2}{4}{VDMMid!49.46!VDMLow} &
 \VDMCell{0.00}{0}{4}{VDMMid!0.00!VDMLow} \\

\VDMTask{P19}{Projection concurrency} & M &
 \VDMCell{67.14}{4}{4}{VDMHigh!34.28!VDMMid} &
 \VDMCell{45.62}{3}{4}{VDMMid!91.24!VDMLow} &
 \VDMCell{48.69}{4}{4}{VDMMid!97.38!VDMLow} &
 \VDMCell{\textbf{69.64}}{4}{4}{VDMHigh!39.28!VDMMid} &
 \VDMCell{32.45}{4}{4}{VDMMid!64.90!VDMLow} &
 \VDMCell{11.62}{3}{4}{VDMMid!23.24!VDMLow} &
 \VDMCell{7.12}{2}{4}{VDMMid!14.24!VDMLow} &
 \VDMCell{3.38}{1}{4}{VDMMid!6.76!VDMLow} \\

\VDMTask{P20}{Refraction and reflection} & H &
 \VDMCell{9.09}{2}{4}{VDMMid!18.18!VDMLow} &
 \VDMCell{31.67}{4}{4}{VDMMid!63.34!VDMLow} &
 \VDMCell{13.69}{3}{4}{VDMMid!27.38!VDMLow} &
 \VDMCell{22.86}{2}{4}{VDMMid!45.72!VDMLow} &
 \VDMCell{14.50}{2}{4}{VDMMid!29.00!VDMLow} &
 \VDMCell{14.88}{3}{4}{VDMMid!29.76!VDMLow} &
 \VDMCell{\textbf{39.66}}{4}{4}{VDMMid!79.32!VDMLow} &
 \VDMCell{7.87}{2}{4}{VDMMid!15.74!VDMLow} \\

\VDMGroup{5. Hydrostatics and Buoyancy}

\VDMTask{P21}{Communicating vessels} & E &
 \VDMCell{97.01}{4}{4}{VDMHigh!94.02!VDMMid} &
 \VDMCell{98.16}{4}{4}{VDMHigh!96.32!VDMMid} &
 \VDMCell{97.70}{4}{4}{VDMHigh!95.40!VDMMid} &
 \VDMCell{\textbf{99.01}}{4}{4}{VDMHigh!98.02!VDMMid} &
 \VDMCell{93.09}{4}{4}{VDMHigh!86.18!VDMMid} &
 \VDMCell{97.40}{4}{4}{VDMHigh!94.80!VDMMid} &
 \VDMCell{95.44}{4}{4}{VDMHigh!90.88!VDMMid} &
 \VDMCell{94.05}{4}{4}{VDMHigh!88.10!VDMMid} \\

\VDMTask{P22}{Floating-ice immersion} & E &
 \VDMCell{66.56}{4}{4}{VDMHigh!33.12!VDMMid} &
 \VDMCell{83.22}{4}{4}{VDMHigh!66.44!VDMMid} &
 \VDMCell{78.82}{4}{4}{VDMHigh!57.64!VDMMid} &
 \VDMCell{85.29}{4}{4}{VDMHigh!70.58!VDMMid} &
 \VDMCell{\textbf{85.86}}{4}{4}{VDMHigh!71.72!VDMMid} &
 \VDMCell{68.96}{4}{4}{VDMHigh!37.92!VDMMid} &
 \VDMCell{67.32}{4}{4}{VDMHigh!34.64!VDMMid} &
 \VDMCell{73.38}{4}{4}{VDMHigh!46.76!VDMMid} \\

\VDMTask{P23}{Liquid-surface orientation} & M &
 \VDMCell{57.52}{3}{4}{VDMHigh!15.04!VDMMid} &
 \VDMCell{\textbf{72.68}}{4}{4}{VDMHigh!45.36!VDMMid} &
 \VDMCell{8.32}{1}{4}{VDMMid!16.64!VDMLow} &
 \VDMCell{0.00}{0}{4}{VDMMid!0.00!VDMLow} &
 \VDMCell{21.06}{2}{4}{VDMMid!42.12!VDMLow} &
 \VDMCell{48.36}{3}{4}{VDMMid!96.72!VDMLow} &
 \VDMCell{34.72}{3}{4}{VDMMid!69.44!VDMLow} &
 \VDMCell{0.00}{0}{4}{VDMMid!0.00!VDMLow} \\

\VDMGroup{6. Phase Transitions and Melting}

\VDMTask{P24}{Freezing-induced expansion} & E &
 \VDMCell{94.30}{4}{4}{VDMHigh!88.60!VDMMid} &
 \VDMCell{\textbf{96.21}}{4}{4}{VDMHigh!92.42!VDMMid} &
 \VDMCell{66.69}{3}{4}{VDMHigh!33.38!VDMMid} &
 \VDMCell{24.38}{1}{4}{VDMMid!48.76!VDMLow} &
 \VDMCell{0.00}{0}{4}{VDMMid!0.00!VDMLow} &
 \VDMCell{86.43}{4}{4}{VDMHigh!72.86!VDMMid} &
 \VDMCell{23.71}{1}{4}{VDMMid!47.42!VDMLow} &
 \VDMCell{0.00}{0}{4}{VDMMid!0.00!VDMLow} \\

\VDMTask{P25}{Ice melting: water level} & H &
 \VDMCell{\textbf{95.61}}{4}{4}{VDMHigh!91.22!VDMMid} &
 \VDMCell{0.00}{0}{4}{VDMMid!0.00!VDMLow} &
 \VDMCell{15.00}{4}{4}{VDMMid!30.00!VDMLow} &
 \VDMCell{0.00}{0}{4}{VDMMid!0.00!VDMLow} &
 \VDMCell{0.00}{0}{4}{VDMMid!0.00!VDMLow} &
 \VDMCell{3.75}{1}{4}{VDMMid!7.50!VDMLow} &
 \VDMCell{0.00}{0}{4}{VDMMid!0.00!VDMLow} &
 \VDMCell{0.00}{0}{4}{VDMMid!0.00!VDMLow} \\

\VDMTask{P26}{Ice with a stone: melting} & H &
 \VDMCell{\textbf{15.00}}{4}{4}{VDMMid!30.00!VDMLow} &
 \VDMCell{0.00}{0}{4}{VDMMid!0.00!VDMLow} &
 \VDMCell{3.75}{1}{4}{VDMMid!7.50!VDMLow} &
 \VDMCell{0.00}{0}{4}{VDMMid!0.00!VDMLow} &
 \VDMCell{0.00}{0}{4}{VDMMid!0.00!VDMLow} &
 \VDMCell{8.81}{2}{4}{VDMMid!17.62!VDMLow} &
 \VDMCell{0.00}{0}{4}{VDMMid!0.00!VDMLow} &
 \VDMCell{0.00}{0}{4}{VDMMid!0.00!VDMLow} \\

\VDMTask{P27}{Freshwater ice in saltwater} & H &
 \VDMCell{46.87}{4}{4}{VDMMid!93.74!VDMLow} &
 \VDMCell{0.00}{0}{4}{VDMMid!0.00!VDMLow} &
 \VDMCell{\textbf{57.50}}{4}{4}{VDMHigh!15.00!VDMMid} &
 \VDMCell{0.00}{0}{4}{VDMMid!0.00!VDMLow} &
 \VDMCell{0.00}{0}{4}{VDMMid!0.00!VDMLow} &
 \VDMCell{0.00}{0}{4}{VDMMid!0.00!VDMLow} &
 \VDMCell{0.00}{0}{4}{VDMMid!0.00!VDMLow} &
 \VDMCell{3.75}{1}{4}{VDMMid!7.50!VDMLow} \\

\VDMTask{P28}{Crushed vs. intact ice} & H &
 \VDMCell{\textbf{57.50}}{4}{4}{VDMHigh!15.00!VDMMid} &
 \VDMCell{36.25}{4}{4}{VDMMid!72.50!VDMLow} &
 \VDMCell{15.00}{4}{4}{VDMMid!30.00!VDMLow} &
 \VDMCell{15.00}{4}{4}{VDMMid!30.00!VDMLow} &
 \VDMCell{15.00}{4}{4}{VDMMid!30.00!VDMLow} &
 \VDMCell{7.50}{2}{4}{VDMMid!15.00!VDMLow} &
 \VDMCell{7.50}{2}{4}{VDMMid!15.00!VDMLow} &
 \VDMCell{0.00}{0}{4}{VDMMid!0.00!VDMLow} \\

\VDMGroup{7. Electrostatics, Magnetism, and Electromagnetic Induction}

\VDMTask{P29}{Eddy-current braking} & E &
 \VDMCell{36.25}{4}{4}{VDMMid!72.50!VDMLow} &
 \VDMCell{\textbf{89.38}}{4}{4}{VDMHigh!78.76!VDMMid} &
 \VDMCell{78.75}{4}{4}{VDMHigh!57.50!VDMMid} &
 \VDMCell{39.38}{2}{4}{VDMMid!78.76!VDMLow} &
 \VDMCell{53.75}{3}{4}{VDMHigh!7.50!VDMMid} &
 \VDMCell{57.50}{4}{4}{VDMHigh!15.00!VDMMid} &
 \VDMCell{32.50}{3}{4}{VDMMid!65.00!VDMLow} &
 \VDMCell{3.75}{1}{4}{VDMMid!7.50!VDMLow} \\

\VDMTask{P30}{Coil-induced light emission} & E &
 \VDMCell{57.50}{4}{4}{VDMHigh!15.00!VDMMid} &
 \VDMCell{57.50}{4}{4}{VDMHigh!15.00!VDMMid} &
 \VDMCell{\textbf{68.12}}{4}{4}{VDMHigh!36.24!VDMMid} &
 \VDMCell{57.50}{4}{4}{VDMHigh!15.00!VDMMid} &
 \VDMCell{57.50}{4}{4}{VDMHigh!15.00!VDMMid} &
 \VDMCell{\textbf{68.12}}{4}{4}{VDMHigh!36.24!VDMMid} &
 \VDMCell{32.50}{3}{4}{VDMMid!65.00!VDMLow} &
 \VDMCell{3.75}{1}{4}{VDMMid!7.50!VDMLow} \\

\VDMTask{P31}{Charged-sphere equilibrium} & M &
 \VDMCell{\textbf{98.34}}{4}{4}{VDMHigh!96.68!VDMMid} &
 \VDMCell{97.24}{4}{4}{VDMHigh!94.48!VDMMid} &
 \VDMCell{15.00}{4}{4}{VDMMid!30.00!VDMLow} &
 \VDMCell{15.00}{4}{4}{VDMMid!30.00!VDMLow} &
 \VDMCell{15.00}{4}{4}{VDMMid!30.00!VDMLow} &
 \VDMCell{10.88}{3}{4}{VDMMid!21.76!VDMLow} &
 \VDMCell{14.81}{4}{4}{VDMMid!29.62!VDMLow} &
 \VDMCell{15.00}{4}{4}{VDMMid!30.00!VDMLow} \\

\VDMTask{P32}{Final compass orientations} & M &
 \VDMCell{\textbf{36.43}}{4}{4}{VDMMid!72.86!VDMLow} &
 \VDMCell{15.05}{4}{4}{VDMMid!30.10!VDMLow} &
 \VDMCell{15.00}{4}{4}{VDMMid!30.00!VDMLow} &
 \VDMCell{18.01}{4}{4}{VDMMid!36.02!VDMLow} &
 \VDMCell{36.32}{4}{4}{VDMMid!72.64!VDMLow} &
 \VDMCell{16.41}{4}{4}{VDMMid!32.82!VDMLow} &
 \VDMCell{15.00}{4}{4}{VDMMid!30.00!VDMLow} &
 \VDMCell{15.95}{4}{4}{VDMMid!31.90!VDMLow} \\

\VDMTask{P33}{Closed vs. open jumping rings} & M &
 \VDMCell{23.00}{4}{4}{VDMMid!46.00!VDMLow} &
 \VDMCell{20.84}{4}{4}{VDMMid!41.68!VDMLow} &
 \VDMCell{15.00}{4}{4}{VDMMid!30.00!VDMLow} &
 \VDMCell{15.00}{4}{4}{VDMMid!30.00!VDMLow} &
 \VDMCell{36.25}{4}{4}{VDMMid!72.50!VDMLow} &
 \VDMCell{\textbf{57.50}}{4}{4}{VDMHigh!15.00!VDMMid} &
 \VDMCell{15.00}{4}{4}{VDMMid!30.00!VDMLow} &
 \VDMCell{15.00}{4}{4}{VDMMid!30.00!VDMLow} \\

\VDMTask{P34}{Solid vs. slotted plate damping} & H &
 \VDMCell{\textbf{36.25}}{4}{4}{VDMMid!72.50!VDMLow} &
 \VDMCell{15.00}{4}{4}{VDMMid!30.00!VDMLow} &
 \VDMCell{15.00}{4}{4}{VDMMid!30.00!VDMLow} &
 \VDMCell{0.00}{0}{4}{VDMMid!0.00!VDMLow} &
 \VDMCell{15.00}{4}{4}{VDMMid!30.00!VDMLow} &
 \VDMCell{7.50}{2}{4}{VDMMid!15.00!VDMLow} &
 \VDMCell{15.00}{4}{4}{VDMMid!30.00!VDMLow} &
 \VDMCell{10.50}{3}{4}{VDMMid!21.00!VDMLow} \\

\VDMGroup{8. Granular Media and Discharge Flow}

\VDMTask{P35}{Sandpile angle scaling} & E &
 \VDMCell{\textbf{99.08}}{4}{4}{VDMHigh!98.16!VDMMid} &
 \VDMCell{94.21}{4}{4}{VDMHigh!88.42!VDMMid} &
 \VDMCell{73.64}{4}{4}{VDMHigh!47.28!VDMMid} &
 \VDMCell{98.55}{4}{4}{VDMHigh!97.10!VDMMid} &
 \VDMCell{98.08}{4}{4}{VDMHigh!96.16!VDMMid} &
 \VDMCell{30.11}{4}{4}{VDMMid!60.22!VDMLow} &
 \VDMCell{97.18}{4}{4}{VDMHigh!94.36!VDMMid} &
 \VDMCell{93.18}{4}{4}{VDMHigh!86.36!VDMMid} \\

\VDMTask{P36}{Sand vs. water discharge} & H &
 \VDMCell{\textbf{38.39}}{4}{4}{VDMMid!76.78!VDMLow} &
 \VDMCell{21.89}{4}{4}{VDMMid!43.78!VDMLow} &
 \VDMCell{15.25}{4}{4}{VDMMid!30.50!VDMLow} &
 \VDMCell{7.12}{2}{4}{VDMMid!14.24!VDMLow} &
 \VDMCell{16.44}{3}{4}{VDMMid!32.88!VDMLow} &
 \VDMCell{10.69}{3}{4}{VDMMid!21.38!VDMLow} &
 \VDMCell{7.88}{2}{4}{VDMMid!15.76!VDMLow} &
 \VDMCell{7.81}{2}{4}{VDMMid!15.62!VDMLow} \\

\VDMGroup{9. Surface Tension and Viscous Flow}

\VDMTask{P37}{Capillary rise vs. diameter} & E &
 \VDMCell{78.75}{4}{4}{VDMHigh!57.50!VDMMid} &
 \VDMCell{\textbf{100.00}}{4}{4}{VDMHigh!100.00!VDMMid} &
 \VDMCell{57.12}{4}{4}{VDMHigh!14.24!VDMMid} &
 \VDMCell{\textbf{100.00}}{4}{4}{VDMHigh!100.00!VDMMid} &
 \VDMCell{36.25}{4}{4}{VDMMid!72.50!VDMLow} &
 \VDMCell{11.44}{3}{4}{VDMMid!22.88!VDMLow} &
 \VDMCell{15.00}{4}{4}{VDMMid!30.00!VDMLow} &
 \VDMCell{10.88}{3}{4}{VDMMid!21.76!VDMLow} \\

\VDMTask{P38}{Viscous settling speed} & E &
 \VDMCell{66.88}{4}{4}{VDMHigh!33.76!VDMMid} &
 \VDMCell{67.25}{4}{4}{VDMHigh!34.50!VDMMid} &
 \VDMCell{68.50}{4}{4}{VDMHigh!37.00!VDMMid} &
 \VDMCell{\textbf{69.85}}{4}{4}{VDMHigh!39.70!VDMMid} &
 \VDMCell{50.86}{3}{4}{VDMHigh!1.72!VDMMid} &
 \VDMCell{65.91}{4}{4}{VDMHigh!31.82!VDMMid} &
 \VDMCell{51.39}{3}{4}{VDMHigh!2.78!VDMMid} &
 \VDMCell{58.97}{4}{4}{VDMHigh!17.94!VDMMid} \\

\VDMTask{P39}{Bubble-film curvature} & M &
 \VDMCell{15.00}{4}{4}{VDMMid!30.00!VDMLow} &
 \VDMCell{\textbf{55.08}}{4}{4}{VDMHigh!10.16!VDMMid} &
 \VDMCell{25.30}{4}{4}{VDMMid!50.60!VDMLow} &
 \VDMCell{28.79}{4}{4}{VDMMid!57.58!VDMLow} &
 \VDMCell{42.26}{4}{4}{VDMMid!84.52!VDMLow} &
 \VDMCell{32.41}{4}{4}{VDMMid!64.82!VDMLow} &
 \VDMCell{42.62}{4}{4}{VDMMid!85.24!VDMLow} &
 \VDMCell{15.00}{4}{4}{VDMMid!30.00!VDMLow} \\

\VDMTask{P40}{Droplet volume conservation} & M &
 \VDMCell{58.30}{4}{4}{VDMHigh!16.60!VDMMid} &
 \VDMCell{43.52}{4}{4}{VDMMid!87.04!VDMLow} &
 \VDMCell{32.46}{4}{4}{VDMMid!64.92!VDMLow} &
 \VDMCell{37.07}{2}{4}{VDMMid!74.14!VDMLow} &
 \VDMCell{26.31}{3}{4}{VDMMid!52.62!VDMLow} &
 \VDMCell{\textbf{62.00}}{4}{4}{VDMHigh!24.00!VDMMid} &
 \VDMCell{14.62}{4}{4}{VDMMid!29.24!VDMLow} &
 \VDMCell{0.00}{0}{4}{VDMMid!0.00!VDMLow} \\

\midrule

\multicolumn{2}{@{}l}{\textbf{Average score}} &
 \VDMAvgCell{\textbf{57.76}}{VDMHigh!15.52!VDMMid} &
 \VDMAvgCell{54.89}{VDMHigh!9.77!VDMMid} &
 \VDMAvgCell{42.46}{VDMMid!85.81!VDMLow} &
 \VDMAvgCell{37.79}{VDMMid!75.58!VDMLow} &
 \VDMAvgCell{35.66}{VDMMid!71.32!VDMLow} &
 \VDMAvgCell{32.25}{VDMMid!64.50!VDMLow} &
 \VDMAvgCell{31.97}{VDMMid!63.94!VDMLow} &
 \VDMAvgCell{18.81}{VDMMid!37.62!VDMLow} \\

\bottomrule
\end{tabular}
\endgroup

\end{minipage}
\end{lrbox}
\ifdim\dimexpr\ht\VDMTableBox+\dp\VDMTableBox\relax>.94\textheight
  \resizebox*{!}{.94\textheight}{\usebox{\VDMTableBox}}%
\else
  \usebox{\VDMTableBox}%
\fi
\end{table*}

\begin{figure}[t]
    \centering
    \includegraphics[width=\linewidth]{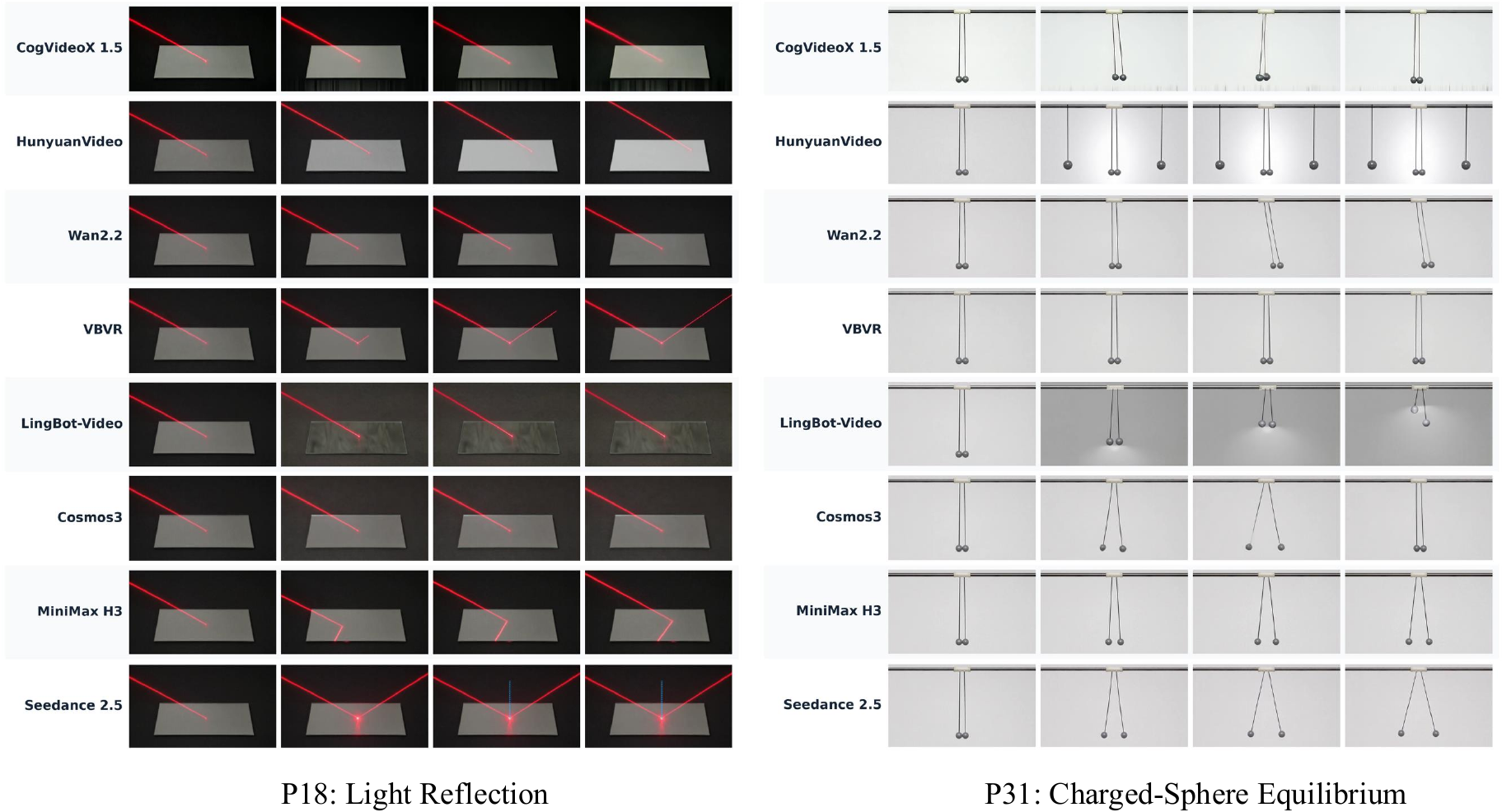}
    \caption{\textbf{Qualitative comparison across eight video generation models.} We show results on P18: Light Reflection (left) and P31: Charged-Sphere Equilibrium (right). Each row corresponds to a model, with four frames displayed in temporal order from left to right within each task. The examples reveal differences in reflected-ray geometry and charged-sphere configurations, with visible failures including missing reflected rays and the appearance of additional spheres.}
    \label{fig:vis_comp}
\end{figure}

\begin{figure}[t]
    \centering
    \includegraphics[width=\linewidth]{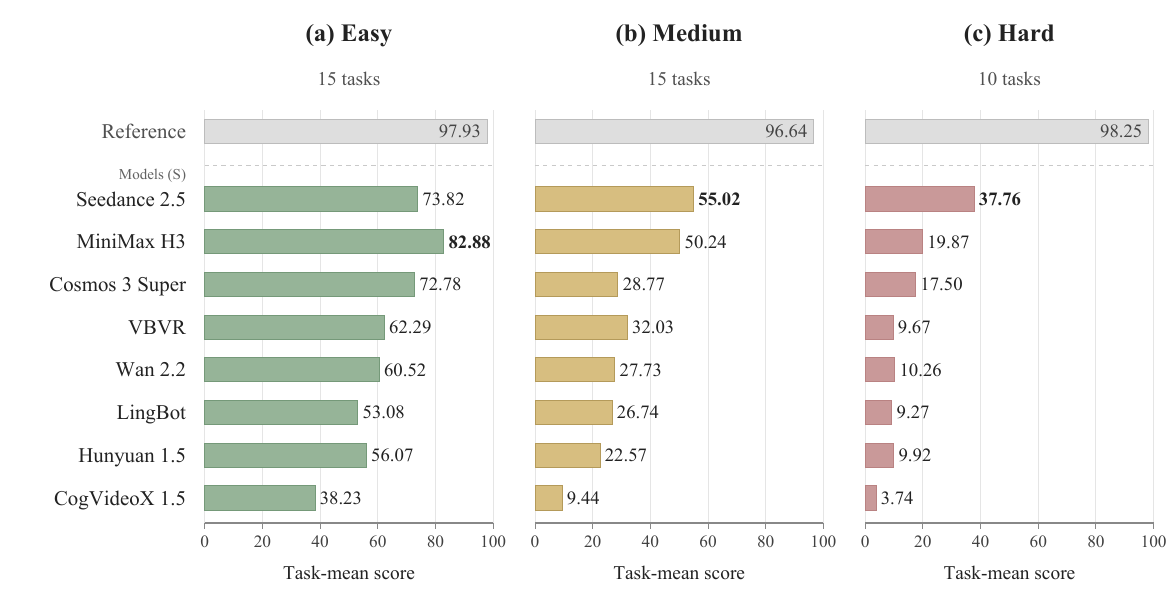}
    \caption{Performance of each video generation model across Easy, Medium, and Hard physical phenomena. Colored bars show composite scores, while light-gray bars show physical scores on synthetic reference videos with the consistency gate bypassed. All scores are scaled to 0--100. The best generation model for each difficulty level is shown in bold.}
    \label{fig:difficulty_comparison}
\end{figure}


\textbf{Results by Task.} Table~\ref{tab:vdm-task-model} shows substantial variation in composite scores across models and tasks under the strict per-video consistency gate ($C \ge 80$). Even the strongest model, Seedance, achieves an average composite score of approximately 57.76, indicating considerable room for improvement in generating videos that both satisfy the intended task and reproduce its physical behavior. Four tasks illustrate this variation. Melting ice containing a stone (P26) remains challenging for every evaluated model, with a model-averaged composite score of 3.45 and a maximum of 15.00. All eight models receive zero physical scores under the evaluation protocol, so their nonzero composite scores arise entirely from the consistency term. Light reflection (P18) reveals pronounced differences between models: Seedance and VBVR achieve scores of 96.15 and 95.75, respectively, whereas all other models score at most 49.82. Figure~\ref{fig:vis_comp} (left) illustrates corresponding differences in the generated sequences, including clearly visible reflected rays in the Seedance and VBVR examples and missing reflections in several other outputs. Charged-sphere equilibrium (P31) exhibits a similarly large performance gap: Seedance and MiniMax score 98.34 and 97.24, respectively, whereas all other models score at most 15.00. As illustrated in Figure~\ref{fig:vis_comp} (right), these two models generate outward-separated sphere configurations, while some other models show little separation or introduce additional spheres. Communicating-vessel equilibrium (P21), however, provides an example of consistently strong performance. All eight models score between 93.09 and 99.01 and pass the automatic consistency gate on every available video, indicating broad success on this particular equilibrium task.

\textbf{Results by Difficulty Level.}
The empirical difficulty groups, defined by mean composite scores across the evaluated models, summarize differences in observed task performance. Easy tasks include communicating-vessel equilibrium (P21), floating-ice immersion (P22), and hanging-chain equilibrium (P7). These examples involve relationships that can be evaluated from relatively stable configurations, which may partly explain their stronger performance. Medium tasks include free fall (P2) and pure rolling (P8), where reproducing recognizable motion is insufficient: the generated trajectories must also satisfy quantitative constraints on fall dynamics or the coupling between translation and rotation. For example, all available free-fall videos pass the automatic consistency gate ($C \ge 80$), but the average composite score across models is only 26.61, showing that passing this gate does not guarantee a high physical score. Hard tasks include equal-mass collisions (P5), rough-incline round trips (P11), and melting ice containing a stone (P26). These scenarios require maintaining physical relationships through interactions or successive events, such as contact, motion reversal, and changes in object configuration. Their low scores may reflect difficulties coordinating these processes over time. Failures at the consistency stage also contribute: only 56.9\% of available Hard videos pass the automatic consistency gate, compared with 89.4\% of Easy videos. Thus, low composite scores can reflect both failures to realize the required task conditions and low physical scores under the measurement protocol. Model strengths also differ across these groups: MiniMax leads on Easy tasks, whereas Seedance achieves the highest task-averaged Hard score of 37.76, substantially above MiniMax's 19.87, although its absolute score remains low.

\textbf{Evaluator Validation on Synthetic Reference Videos.}
We assess whether the evaluator assigns appropriately high physical scores to synthetic reference videos designed to satisfy prescribed physical relationships. These videos are generated as 2D animations using existing tools: object states are computed from the specified relations and rendered frame by frame, while appearance variants preserve the underlying trajectories. We evaluate 480 reference videos across 40 tasks, with 12 videos per task. To isolate the physical measurement module, we bypass the VLM consistency gate. As shown by the light-gray bars in Figure~\ref{fig:difficulty_comparison}, the reference videos receive mean physical scores of 97.93, 96.64, and 98.25 on Easy, Medium, and Hard tasks, respectively, with an overall mean of 97.53 out of 100. These results support the evaluator's agreement with the prescribed physical relationships under controlled synthetic conditions.

\subsection{Agreement with Human Judgments}

To assess how well automatic physical scores align with human judgments, we compare our evaluator with a direct scoring baseline based on Qwen3.6-27B~\citep{yang2025qwen3}, denoted as Direct VLM. Given 24 sampled frames, the original generation prompt, and the reference image, Direct VLM directly predicts a physical correctness score in $[0,1]$. Frames are resized to a maximum side length of 640 pixels, and inference uses a temperature of zero. Both methods are evaluated using human annotations for 320 videos across 40 tasks. Ten annotators participate in the human evaluation. We additionally report Human--Human agreement as a descriptive reference for inter-annotator consistency.

\begin{figure}[t]
    \centering
    \includegraphics[width=\linewidth]{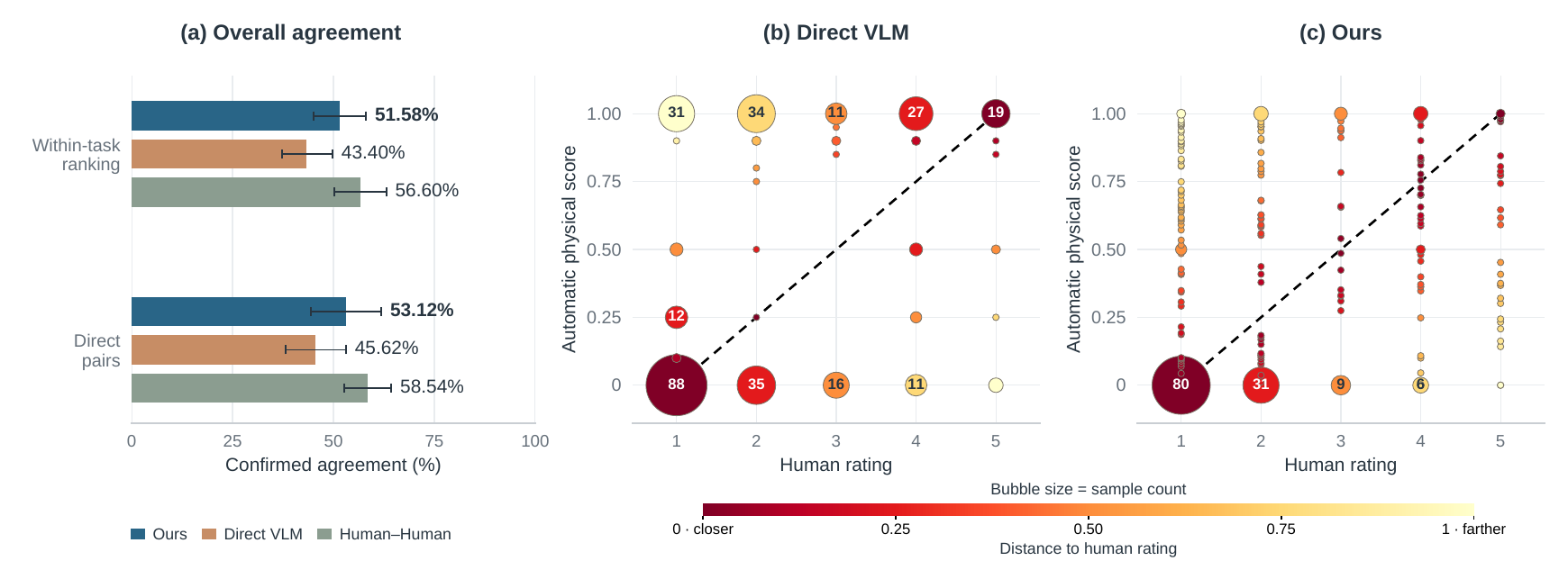}
    \caption{Agreement with human judgments. (a) Within-task ranking agreement and confirmed pairwise agreement for our evaluator, Direct VLM, and the Human--Human reference. Matching ties count as agreement. (b,c) Automatic physical scores versus the five-level reference human ratings for Direct VLM and our evaluator, respectively. Bubble areas indicate video counts, with selected counts labeled. Black dashed lines show the illustrative reference $P=(H-1)/4$. Colors indicate absolute deviations from this reference, with darker colors representing smaller deviations.}
    \label{fig:human}
\end{figure}

\textbf{Within-task Ranking Agreement.} Higher automatic physical scores should correspond to greater human-perceived physical correctness.  Annotators rate individual videos on a five-point scale, ranging from clear physical violation to no obvious violation, with a separate option for insufficient evidence. Let $H_i\in\{1,2,3,4,5\}$ denote the reference annotator's rating for video $i$. Both automatic methods are compared against these same reference ratings, while Human--Human agreement compares the orderings induced by the annotators' ratings.

The 320 videos across 40 tasks yield $40\times\binom{8}{2}=1{,}120$ within-task video pairs. We exclude 203 pairs involving a video that at least one annotator could not assess, leaving a common set $\mathcal C$ of 917 pairs for all three comparisons. Human and automatic ties are retained. We pool these pairs across tasks and compute:
\begin{equation}
    A_{\mathrm{rank}}
    =\frac{1}{|\mathcal C|}
    \sum_{(i,j)\in\mathcal C}
    \mathbf{1}\!\left[
        \operatorname{sign}(P_i-P_j)
        =\operatorname{sign}(H_i-H_j)
    \right],
    \label{eq:human_ranking_agreement}
\end{equation}
where $P_i$ denotes the automatic physical score and $\operatorname{sign}(0)=0$. Each video pair is counted once, and agreement requires the same strict ordering or a tie from both sources. Human--Human agreement applies the same rule to the annotators' ratings over the same 917 pairs.

As shown in Figure~\ref{fig:human}(a), our evaluator agrees with the reference human ordering on 473 pairs (51.58\%), compared with 398 pairs (43.40\%) for Direct VLM and 519 pairs (56.60\%) for Human--Human. Our evaluator therefore improves over Direct VLM by 8.18 percentage points and falls 5.02 percentage points below the Human--Human reference. Figure~\ref{fig:human}(b,c) shows the corresponding individual-video score distributions using the same reference ratings. The mapping $P=(H-1)/4$ is used solely for visualization and does not enter either agreement metric.

\textbf{Confirmed Pairwise Agreement.} Annotators compare 160 predefined within-task video pairs and select video A, video B, a tie, or insufficient evidence. Human preferences are determined by a strict majority of assessable votes, with insufficient-evidence responses treated as abstentions. Automatic preferences follow the ordering of the two physical scores, including ties. We compute confirmed agreement over all 160 pairs:
\begin{equation}
    A_{\mathrm{pair}}^{\mathrm{conf}}
    =\frac{N_{\mathrm{agree}}}{160},
    \label{eq:human_pairwise_agreement}
\end{equation}
where $N_{\mathrm{agree}}$ counts pairs with matching human and automatic A/B/tie judgments. Pairs without a majority human judgment remain in the denominator but do not count as agreements. The Human--Human reference averages exact A/B/tie agreement across the three annotator pairings, retaining all 160 video pairs in each denominator and counting only matching assessable responses as agreement.

As shown in Figure~\ref{fig:human}(a), our evaluator achieves 85 confirmed agreements (53.12\%), compared with 73 (45.62\%) for Direct VLM and 58.54\% for the Human--Human reference. This represents an improvement of 7.50 percentage points over Direct VLM and a gap of 5.42 percentage points to Human--Human agreement. Across both measures, our evaluator achieves higher agreement with human judgments than Direct VLM, while remaining below the Human--Human reference. Error bars indicate 95\% confidence intervals at the task level.


\section{Related Work}
\label{sec:related_work}


\paragraph{Physics-Aware Video Generation.}
Physics-aware video generation incorporates physical structure through simulation, reasoning, representation learning, and post-training.
Simulation-based approaches combine rigid-body, deformable-body, or continuum dynamics with diffusion-based synthesis~\citep{liu2024physgen,xie2025physanimator,montanaro2024motioncraft,tan2026physmotion}.
PhysCtrl~\citep{wang2025physctrl} guides video synthesis with generated 3D trajectories, while PhysDreamer~\citep{zhang2024physdreamer} transfers video dynamics priors to interactive 3D objects.
Reasoning-guided methods incorporate physical context through prompt refinement and explicit planning~\citep{xue2025phyt2v,zhang2025think,zhao2026phyrpr,feng2026newton,yang2025vlipp}.
At the representation level, WISA~\citep{wang2025wisa} encodes physical descriptions and properties, while VideoREPA~\citep{zhang2026videorepa} aligns spatiotemporal token relations with video foundation models.
Recent approaches further incorporate retrieved examples, joint RGB-perception modeling, and latent physical dynamics~\citep{cheng2026physrag,lin2026mmphysvideo,shen2026phantom}.
Post-training methods explore AI feedback, task-specific rewards, and learned physical representations to improve generated dynamics~\citep{furuta2024improving,li2025pisa,ji2025physmaster,zhang2026physrvg}.
NewtonRewards~\citep{le2025newtonrewards} constructs physics rewards from measurable proxies, while PhyGDPO and PhyWorld optimize generation using physics-aware preferences~\citep{cai2026phygdpo,zhao2026phyworld}.
These advances motivate task-specific tests that determine which physical relationships generated videos preserve under stated assumptions.
Our benchmark provides such tests through controlled experimental conditions and explicit quantitative measurements.

\paragraph{Video Generation Evaluation.}
Existing benchmarks use four complementary forms of evidence: general quality metrics, evaluator judgments, reference observations, and direct physical measurements.
General quality benchmarks assess visual fidelity, temporal coherence, and compositional alignment~\citep{huang2024vbench,liu2024evalcrafter,sun2025t2v}, but these criteria alone do not establish physical correctness.
Judgment-based benchmarks assess physics and commonsense through human annotations, VLMs, or learned evaluators~\citep{zheng2025vbench,bansal2025videophy,bansal2026videophy2,meng2025phygenbench,guo2025t2vphysbench,gu2026phyworldbench,li2026worldmodelbench}, with finer-grained criteria and structured questions improving interpretability~\citep{lin2026phyground,pothiraj2026pqsg}.
However, their scores reflect evaluator judgments rather than directly measured deviations in physical quantities.
Reference-based approaches, including Physics-IQ, Physics-IQ Verified, and RigidBench, compare generations with real or simulated observations~\citep{motamed2025generativevideomodelsunderstand,radsch2026physics,jain2026rigidbench}.
They require suitable reference data, and reference similarity alone does not isolate violations of specific physical laws.
Measurement-based approaches provide more explicit diagnostics: PhysWeep estimates physical parameters~\citep{khanbayov2026physweep}, Principia tests paired-object relationships~\citep{thozhiyoor2026principia}, and GAUGE combines measured trajectories with calibrated physical metadata~\citep{wang2026gauge}.
Their video evaluations primarily target selected mechanical dynamics, leaving room for quantitative tests of other physical phenomena.
We complement these approaches with controlled experiments spanning mechanics, optics, and other physical domains.
It quantifies deviations from task-specific spatial, temporal, and spatiotemporal relationships under stated assumptions, using scale-canceling comparisons where applicable and computing physical scores from measurements without VLM judgments of physical correctness.
\section{Conclusion}
We introduced \textit{World Models' Last Exam in Physics}, a benchmark for evaluating physical consistency in video generation through 40 controlled tasks and explicit, task-specific measurements. Our evaluation framework separates temporal consistency and task observability from physical correctness, grounding scores in measurable relationships and retaining evidence for interpreting failures. Experiments across eight video generation models reveal persistent difficulties in reproducing quantitative physical relationships and completing the required physical events. Evaluation on synthetic reference videos supports the measurement module under controlled conditions, while human evaluation shows higher confirmed agreement than direct VLM scoring in both within-task rankings and pairwise comparisons. These findings highlight the value of explicit physical measurements for assessing generated videos. Current evaluation remains constrained by the visibility and reliable extraction of task-relevant evidence. Future work will extend task coverage and improve measurement robustness in more complex scenes. We hope this benchmark provides a useful foundation for diagnosing physical inconsistencies and developing video generation models that more reliably reproduce physical phenomena.


\clearpage
\newpage













\clearpage
\newpage
\bibliographystyle{assets/plainnat}
\bibliography{ref}


\clearpage
\newpage
\appendix


\section{Case Study on Evaluator}
\label{sec:case_study_evaluator}

\begin{figure}[t]
    \centering
    \includegraphics[width=0.8\linewidth]{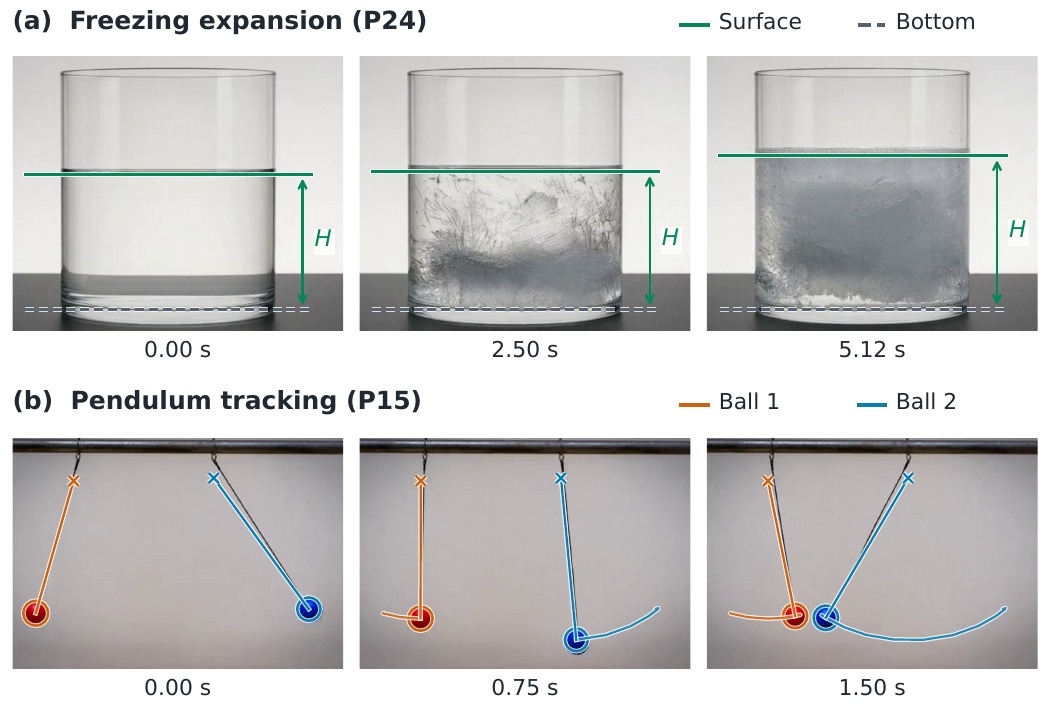}
    \caption{\textbf{Examples of measurement extraction by our evaluator.}
    (a) Freezing expansion (P24): the detected material surface (solid green line) and container bottom (dashed gray line) define the sample height $H$ at three timestamps.
    (b) Pendulum tracking (P15): orange and blue overlays identify the two pendulum bobs and their tracked trajectories.
    The overlays expose the visual measurements used for subsequent physical assessment.}
    \label{fig:case_study_evaluator}
\end{figure}

Figure~\ref{fig:case_study_evaluator} illustrates how our evaluator extracts measurable evidence for task-specific physical assessment.
For freezing expansion (P24), the evaluator identifies the material surface and container bottom to estimate the sample height $H$, enabling assessment of height changes during freezing.
For pendulum tracking (P15), it tracks the two pendulum bobs over time to recover their motion trajectories, providing the basis for quantitative analysis of oscillatory behavior.
These examples demonstrate how intermediate measurements connect generated video content to physical evaluation, while making potential boundary-detection and tracking errors available for inspection.

\section{Detailed Task Catalog}
\label{app:task_catalog}

Our benchmark comprises 40 tasks across nine physical families, with 65 defined evaluation metrics: one primary metric (M1) for every task and a secondary metric (M2) for 25 tasks. Tasks are numbered consecutively from P1 to P40 and grouped by physical family, with Easy tasks followed by Medium and Hard tasks within each family. Table~\ref{tab:task_catalog_overview} summarizes the resulting distribution. Each entry describes the intended scene, the visible quantities measured by the evaluator, and their score normalization.

The difficulty labels follow the current automatic evaluation: composite scores are averaged over available seeds within each model and then equally over the eight models. Ranking tasks by these scores yields 15 Easy, 15 Medium, and 10 Hard tasks. Missing videos are excluded from these averages. These labels describe empirical difficulty for the evaluated models.

\begin{table*}[!htbp]
\centering
\small
\setlength{\tabcolsep}{3.5pt}
\renewcommand{\arraystretch}{1.18}
\caption{The nine physical families, their current task identifiers, evaluation targets, and empirical difficulty distribution. E, M, and H denote Easy, Medium, and Hard.}
\label{tab:task_catalog_overview}
\begin{tabularx}{\textwidth}{@{}>{\raggedright\arraybackslash}p{.25\textwidth}
>{\raggedright\arraybackslash}p{.13\textwidth}Xrrr@{}}
\toprule
Physical family & Task IDs & Physical targets & E & M & H \\
\midrule
Translational motion and collisions
 & P1--P5
 & Restitution, acceleration, projectile geometry, and momentum. & 1 & 2 & 2 \\
Rolling, friction, and rigid-body statics
 & P6--P11
 & Sliding onset, chain geometry, rolling, and supported rotation. & 2 & 3 & 1 \\
Pendulum motion and oscillations
 & P12--P15
 & Period relations for mass, amplitude, and length. & 3 & 1 & 0 \\
Optics and projective geometry
 & P16--P20
 & Projective invariance, ray geometry, and shadow concurrence. & 1 & 3 & 1 \\
Hydrostatics and buoyancy
 & P21--P23
 & Equilibrium levels, submerged fraction, and surface orientation. & 2 & 1 & 0 \\
Phase transitions and melting
 & P24--P28
 & Freezing expansion, melting-induced level changes, and comparative melting evidence. & 1 & 0 & 4 \\
Electrostatics, magnetism, and electromagnetic induction
 & P29--P34
 & Magnetic braking, induction events, symmetry, and needle directions. & 2 & 3 & 1 \\
Granular media and discharge flow
 & P35--P36
 & Repose-angle consistency and flow dependence on filling height. & 1 & 0 & 1 \\
Surface tension and viscous flow
 & P37--P40
 & Capillary ordering, settling speed, film curvature, and droplet volume. & 2 & 2 & 0 \\
\midrule
\textbf{Total} & \textbf{40 tasks} & \textbf{9 physical families} & \textbf{15} & \textbf{15} & \textbf{10} \\
\bottomrule
\end{tabularx}
\end{table*}

\paragraph{Notation and normalization.}
We denote metric residuals by $e_j$ and normalized metric scores by $s_j\in[0,1]$, with higher scores indicating better agreement. Unless a unit is specified, residuals are dimensionless. We use
\[
q(e;a)=\frac{1}{1+|e|/a},\qquad
G(s_1,\ldots,s_n)=\left(\prod_{j=1}^{n}s_j\right)^{1/n},
\]
\[
\operatorname{CV}(x)=\frac{\operatorname{std}(x)}{|\operatorname{mean}(x)|},
\qquad \operatorname{clip}(x)=\min(1,\max(0,x)).
\]
The scale $a$ gives $q=0.5$ at $|e|=a$; it is not a pass threshold. The indicator $\mathbb{I}[A]$ is one when $A$ holds and zero otherwise. Image $y$ increases downward, with the sign of physical height changes specified in each task.

For a task with $m$ defined metrics, the physical score is
\[
P = \frac{100}{m}\sum_{j=1}^{m}s_j, \qquad s_j\in[0,1].
\]
Thus, $P=100s_1$ when M2 is not defined. Consistent with the main evaluation protocol, $C$, $P$, and $S$ are expressed on a 0--100 scale, while individual metric scores remain in $[0,1]$. The automatic composite score used for difficulty assignment follows Eq.~(7):
\[
S = 0.15C + 0.85\widetilde{P}, \qquad
\widetilde{P} =
\begin{cases}
P, & C\geq80 \text{ and a physical score is recorded},\\
0, & \text{otherwise}.
\end{cases}
\]
Here, $C\in[0,100]$ is the automatic consistency and observability score. The metric scores below contribute to $P$; $C$ contributes once per video.

Metrics not defined for a task are excluded from its metric set and averaging denominator. For a task-defined metric, unless a task-specific rule applies, a completed extraction that cannot resolve the required measurement contributes an operational score of zero, while the raw measurement remains undefined. Such an operational zero does not count as a successful measurement or establish a measured physical violation. Input or execution errors do not yield valid physical scores and are excluded from physical-score averages; the treatment of an unavailable physical score in the composite score follows Eq. (7). Each score evaluates only the stated observable relationship, and broader physical properties described by the task remain unverified unless separately measured.

\subsection{Translational Motion and Collisions}
\label{app:catalog_translation}

\paragraph{\normalfont\bfseries P1: Successive bounce heights (Easy).}\mbox{}\par\nobreak
\noindent\textit{Definition.} A ball is dropped onto a horizontal hard surface and produces at least four visible rebounds with decreasing peak heights.

\textit{Metrics.} For adjacent rebounds, define $r_h=h_{\rm next}/h_{\rm prev}$, $e_h^\ast=\sqrt{r_h}$, and $e_t^\ast=\Delta t_{\rm next}/\Delta t_{\rm prev}$. M1 compares height- and timing-based restitution and penalizes energy gain:
\[
e_1=\operatorname{mean}|e_h^\ast-e_t^\ast|,\qquad
e_p=\operatorname{mean}\max(0,e_h^\ast-1).
\]
M2 measures height increase $e_2=\operatorname{mean}\max(0,r_h-1)$ and, when measurable, restitution variability $c=\operatorname{CV}(e_{\rm restitution})$.

\textit{Normalization.} $s_1=G(q(e_1;0.18),q(e_p;0.15))$. M2 is $s_2=G(q(e_2;0.15),q(c;0.35))$ when $c$ is measurable, and $s_2=q(e_2;0.15)$ otherwise. Trackable single motion without repeated rebounds receives $s_1=s_2=0.1$.

\paragraph{\normalfont\bfseries P2: Free fall (Medium).}\mbox{}\par\nobreak
\noindent\textit{Definition.} A ball is released from rest in a fixed side view and falls under gravity before ground contact.

\textit{Metrics.} M1 measures acceleration consistency, $e_1=\operatorname{CV}(\Delta v_y)$, using equal-interval velocity increments. M2 compares displacements $d_1,d_2,d_3$ in three equal intervals:
\[
e_2=\sqrt{\frac{(d_2/(3d_1)-1)^2+(d_3/(5d_1)-1)^2}{2}}.
\]
The first displacement must exceed localization noise; resolved constant-speed descent fails the acceleration test.

\textit{Normalization.} $s_1=q(e_1;0.10)$ and $s_2=q(e_2;0.10)$.

\paragraph{\normalfont\bfseries P3: Complementary-angle projectiles (Medium).}\mbox{}\par\nobreak
\noindent\textit{Definition.} Identical balls are launched simultaneously at $30^\circ$ and $60^\circ$ with equal initial speed, following separate lanes and landing at their launch heights.

\textit{Metrics.} M1 measures range agreement, $e_1=R_{30}/R_{60}-1$. M2 combines the two parabola-fit RMS errors divided by ball diameter, $u_{30},u_{60}$, and initial-speed mismatch $e_v=|v_{30}/v_{60}-1|$.

\textit{Normalization.}
\[
s_1=q(e_1;0.15),\qquad
s_2=G\!\left(q(u_{30};0.30),q(u_{60};0.30),q(e_v;0.20)\right).
\]

\paragraph{\normalfont\bfseries P4: Oblique projectile motion (Hard).}\mbox{}\par\nobreak
\noindent\textit{Definition.} A ball is launched at $45^\circ$, rises to its apex, and returns to its launch height in a fixed side view.

\textit{Metrics.} M1 measures $e_1=H/R-1/4$, where $H$ is apex height and $R$ is range. M2 measures $\operatorname{CV}(v_x)$, $\operatorname{CV}(\Delta v_y)$, and parabola-fit error $u=\operatorname{RMS}/R$. If return to launch height is absent but launch, apex, and descent are resolved, M1 uses $H/(2X_{\rm apex})-1/4$ as a partial-arc estimate.

\textit{Normalization.}
\[
s_1=q(e_1;0.10),\qquad
s_2=G\!\left(q(\operatorname{CV}(v_x);0.10),
q(\operatorname{CV}(\Delta v_y);0.10),q(u;0.05)\right).
\]

\paragraph{\normalfont\bfseries P5: Equal-mass head-on collision (Hard).}\mbox{}\par\nobreak
\noindent\textit{Definition.} A moving red ball strikes an initially stationary blue ball of equal mass on a level surface in a head-on elastic collision.

\textit{Metrics.} M1 measures horizontal momentum agreement:
\[
e_1=\frac{v_{1,\rm after}+v_{2,\rm after}}{v_{1,\rm before}}-1.
\]
M2 is not defined; elasticity is not separately scored.

\textit{Normalization.} $s_1=q(e_1;0.10)$.

\subsection{Rolling, Friction, and Rigid-Body Statics}
\label{app:catalog_rigid_body}

\paragraph{\normalfont\bfseries P6: Sliding onset independent of block mass (Easy).}\mbox{}\par\nobreak
\noindent\textit{Definition.} A hinged board is gradually tilted until two blocks of different masses and matched contact conditions begin sliding.

\textit{Metrics.} M1 compares onset frames, $e_1=|f_l-f_r|/N_f$, where $N_f$ is video length in frames. M2 compares onset inclinations, $e_2=|\theta_{l,\rm onset}-\theta_{r,\rm onset}|$, in degrees.

\textit{Normalization.} $s_1=q(e_1;1.0)$ and $s_2=q(e_2;90^\circ)$. Two measurable blocks that both remain stationary receive $s_1=s_2=0.7$.

\paragraph{\normalfont\bfseries P7: Static shape of a hanging chain (Easy).}\mbox{}\par\nobreak
\noindent\textit{Definition.} A chain with fixed equal-height endpoints is released from a nonequilibrium shape and settles under gravity.

\textit{Metrics.} M1 measures catenary-fit error, $e_1=\operatorname{RMSE}_{\rm contour}/\text{measured sag}$. M2 measures support-height difference divided by horizontal support span, $e_2=\Delta h_{\rm support}/L_{\rm span}$.

\textit{Normalization.} $s_1=q(e_1;0.10)$ and $s_2=q(e_2;0.02)$.

\paragraph{\normalfont\bfseries P8: Pure rolling of a solid sphere (Medium).}\mbox{}\par\nobreak
\noindent\textit{Definition.} A marked sphere is released without a push onto an incline and its horizontal extension, rolling while maintaining surface contact.

\textit{Metrics.} M1 robustly fits center displacement against unwrapped rotation angle, yielding slope $b$ and residual $e_1=|b|/r-1$. M2 measures
\[
e_2=\operatorname{median}\frac{|v-\omega r|}{\max(v,\epsilon)}
\]
over valid moving frames, where $r$ is sphere radius.

\textit{Normalization.} $s_1=q(e_1;0.25)$ and $s_2=q(e_2;0.35)$.

\paragraph{\normalfont\bfseries P9: Rolling solid sphere versus thin ring (Medium).}\mbox{}\par\nobreak
\noindent\textit{Definition.} A solid sphere and a thin ring of equal outer radius roll without slipping from rest at the same height on one incline.

\textit{Metrics.} M1 takes the median ring-to-sphere travel-time ratio $r_t$ over common distances:
\[
e_1=\left|\frac{r_t}{\sqrt{10/7}}-1\right|.
\]
M2 is not defined.

\textit{Normalization.} $s_1=q(e_1;0.10)$.

\paragraph{\normalfont\bfseries P10: Forced rotation about a support edge (Medium).}\mbox{}\par\nobreak
\noindent\textit{Definition.} An actuator pushes a block's upper left face until it rotates about its lower right support edge and lies down without sliding away.

\textit{Metrics.} M1 uses $E=\max(e_{\rm pivot},e_{\rm contact},e_{\rm shape})$. Pivot drift and ground gap are their 95th-percentile values minus twice the pixel noise floor, clipped below at zero and divided by the initial block diagonal. Shape error is noise-adjusted relative edge-length change. The observed rotation $\Delta\theta$ is the largest across continuous intervals of at least four frames and $0.3\,\mathrm{s}$, without interpolating hidden poses. M2 is not defined.

\textit{Normalization.} $s_1=\mathbb{I}[\Delta\theta\geq5^\circ]q(E;0.02)$.

\paragraph{\normalfont\bfseries P11: Sliding up and down a rough incline (Hard).}\mbox{}\par\nobreak
\noindent\textit{Definition.} A block slides up an approximately $30^\circ$ incline, stops, and slides down, with kinetic-friction coefficient $0.20$.

\textit{Metrics.} M1 compares ascent and descent acceleration magnitudes using the measured incline angle $\theta$:
\[
R_\ast=\frac{\sin\theta+0.20\cos\theta}{\sin\theta-0.20\cos\theta},
\qquad e_1=\left|\frac{a_{\rm up}/a_{\rm down}}{R_\ast}-1\right|.
\]
M2 is not defined.

\textit{Normalization.} $s_1=q(e_1;0.10)$. A confirmed ascent-only event receives $s_1=0.1$.

\subsection{Pendulum Motion and Oscillations}
\label{app:catalog_pendulum}

\paragraph{\normalfont\bfseries P12: Pendulum period independent of mass (Easy).}\mbox{}\par\nobreak
\noindent\textit{Definition.} Equal-length pendulums with different bob masses are released from the same angle at the same time.

\textit{Metrics.} M1 compares periods, $e_1=T_{\rm heavy}/T_{\rm light}-1$. M2 compares first-frame lengths using pivots recovered from the swing trajectories:
\[
e_2=\frac{|L_l-L_r|}{\operatorname{mean}(L_l,L_r)}.
\]

\textit{Normalization.} $s_1=q(e_1;0.10)$ and $s_2=q(e_2;0.10)$.

\paragraph{\normalfont\bfseries P13: Finite-amplitude pendulum periods (Easy).}\mbox{}\par\nobreak
\noindent\textit{Definition.} Equal-length pendulums are released from $15^\circ$ and $30^\circ$; the larger-amplitude pendulum should have a longer period.

\textit{Metrics.} Both metrics use $r=T_{\rm large}/T_{\rm small}$. M1 measures $e_1=|r-r_\ast|$; M2 measures the increase $r-1$, where
\[
r_\ast=\frac{K(\sin(30^\circ/2))}{K(\sin(15^\circ/2))}
\simeq1.0130520869
\]
and $K$ is the complete elliptic integral using the modulus convention.

\textit{Normalization.} $s_1=q(e_1;0.05)$ and $s_2=\operatorname{clip}((r-1)/(r_\ast-1))$.

\paragraph{\normalfont\bfseries P14: Pendulum period versus length (Easy).}\mbox{}\par\nobreak
\noindent\textit{Definition.} Identical pendulum bobs on strings with length ratio $1:2$ are released from approximately equal small angles.

\textit{Metrics.} M1 measures
\[
e_1=\left|\frac{(T_{\rm short}/T_{\rm long})^2}{L_{\rm short}/L_{\rm long}}-1\right|.
\]
M2 measures cycle-to-cycle period variability, $c_s=\operatorname{CV}(T_{\rm short})$ and $c_l=\operatorname{CV}(T_{\rm long})$, over repeated oscillations.

\textit{Normalization.} $s_1=q(e_1;0.20)$ and $s_2=G(q(c_s;0.20),q(c_l;0.20))$.

\paragraph{\normalfont\bfseries P15: Small-angle pendulum isochronism (Medium).}\mbox{}\par\nobreak
\noindent\textit{Definition.} Two equal-length pendulums are released simultaneously from different small amplitudes and swing repeatedly.

\textit{Metrics.} M1 compares periods, $e_1=T_l/T_r-1$. M2 fits a sinusoid to each angular trajectory and measures RMS residuals $u_l,u_r$ in radians.

\textit{Normalization.} $s_1=q(e_1;0.10)$ and $s_2=G(q(u_l;0.10\,\mathrm{rad}),q(u_r;0.10\,\mathrm{rad}))$.

\subsection{Optics and Projective Geometry}
\label{app:catalog_optics}

\paragraph{\normalfont\bfseries P16: Cross-ratio of four points on a rigid rod (Easy).}\mbox{}\par\nobreak
\noindent\textit{Definition.} A rod with four fixed collinear markers slides with one endpoint on a wall and the other on the floor.

\textit{Metrics.} For ordered projected marker positions $t_1,\ldots,t_4$, M1 measures $e_1=\operatorname{CV}(\chi)$, where
\[
\chi=\frac{(t_3-t_1)(t_4-t_2)}{(t_3-t_2)(t_4-t_1)}.
\]
M2 measures $e_2=\operatorname{mean}(\operatorname{RMS}_{\rm collinearity}/\text{marker span})$ over frames.

\textit{Normalization.} $s_1=q(e_1;0.10)$ and $s_2=q(e_2;0.05)$.

\paragraph{\normalfont\bfseries P17: Refraction at an air--water interface (Medium).}\mbox{}\par\nobreak
\noindent\textit{Definition.} A fixed laser illuminates a stationary air--water interface, with both incident and refracted rays visible.

\textit{Metrics.} M1 measures Snell's-law residual $e_1=\sin\theta_i/\sin\theta_r-1.333$, with angles relative to the normal. M2 compares ray--interface intersections $p_i,p_r$: $e_2=\|p_i-p_r\|/W_c$, where $W_c$ is calibrated container width.

\textit{Normalization.} $s_1=q(e_1;0.12)$ and $s_2=q(e_2;0.025)$.

\paragraph{\normalfont\bfseries P18: Specular reflection (Medium).}\mbox{}\par\nobreak
\noindent\textit{Definition.} A narrow laser beam strikes a plane mirror obliquely, with both rays and their contact point visible.

\textit{Metrics.} M1 measures $e_1=|\theta_i-\theta_r|$ in degrees. M2 measures $e_2=\operatorname{median}(\text{reflected-ray contact error})/D_f$, where $D_f$ is frame diagonal.

\textit{Normalization.} $s_1=q(e_1;90^\circ)$ and $s_2=q(e_2;1.0)$.

\paragraph{\normalfont\bfseries P19: Concurrent shadows from a point source (Medium).}\mbox{}\par\nobreak
\noindent\textit{Definition.} One stationary point light illuminates four vertical rods; backward extensions of their shadow axes should meet at a common point.

\textit{Metrics.} M1 measures $e_1=\operatorname{RMS}_{\rm concurrence}/D_f$, using shadow-axis line distances to their common intersection and frame diagonal $D_f$. M2 uses pairwise intersection estimates $(x_k,y_k)$:
\[
e_2=\frac{\operatorname{mean}(\operatorname{std}(x_k),\operatorname{std}(y_k))}{D_f}.
\]

\textit{Normalization.} $s_1=q(e_1;0.05)$ and $s_2=q(e_2;0.05)$.

\paragraph{\normalfont\bfseries P20: Refraction and total internal reflection (Hard).}\mbox{}\par\nobreak
\noindent\textit{Definition.} Four distinguishable beams change incidence directions at fixed points on a water interface, exhibiting refraction or total internal reflection.

\textit{Metrics.} M1 measures the consistency of independently inferred refractive indices:
\[
e_1=\frac{\operatorname{std}(n)}{\operatorname{mean}(n)}.
\]
Total internal reflection supplies a bound, not an exact critical-angle measurement. M2 is not defined.

\textit{Normalization.} $s_1=q(e_1;0.05)$.

\subsection{Hydrostatics and Buoyancy}
\label{app:catalog_hydrostatics}

\paragraph{\normalfont\bfseries P21: Liquid levels in communicating vessels (Easy).}\mbox{}\par\nobreak
\noindent\textit{Definition.} Liquid in a U-tube with unequal arm diameters settles after a disturbance, with no liquid added or removed.

\textit{Metrics.} M1 measures final level difference, $e_1=(\operatorname{median}(y_l)-\operatorname{median}(y_r))/H_{\rm ref}$, where $H_{\rm ref}$ is the larger of frame height and the combined arm-region vertical span. M2 measures residual motion, $e_2=\max(|b_l|,|b_r|)/H_{\rm frame}$, using surface-position slopes against frame index. At least five overlapping terminal frames are required.

\textit{Normalization.} $s_1=q(e_1;1.0)$ and $s_2=q(e_2;1\,\mathrm{frame}^{-1})$.

\paragraph{\normalfont\bfseries P22: Submerged fraction of a floating ice column (Easy).}\mbox{}\par\nobreak
\noindent\textit{Definition.} An intact vertical freshwater ice column reaches floating equilibrium without melting or touching its vessel.

\textit{Metrics.} M1 measures the settled submerged-height ratio; M2 compares the left and right waterlines:
\[
e_1=\left|\operatorname{median}\!\left(\frac{h_{\rm sub}}{h_{\rm total}}\right)-0.917\right|,
\qquad e_2=\operatorname{median}\frac{|y_l-y_r|}{W_{\rm vessel}}.
\]

\textit{Normalization.} $s_1=q(e_1;0.10)$ and $s_2=q(e_2;0.05)$.

\paragraph{\normalfont\bfseries P23: Liquid surface perpendicular to gravity (Medium).}\mbox{}\par\nobreak
\noindent\textit{Definition.} A steel ball falls freely beside a vessel with a stationary liquid surface, providing a visible estimate of gravity direction.

\textit{Metrics.} M1 measures $e_1=\angle(\boldsymbol{a}_{\rm ball},\boldsymbol{t}_{\rm surface})-90^\circ$. M2 measures surface line-fit RMS $u_s$ and constant-acceleration trajectory-fit RMS $u_b$, both in pixels.

\textit{Normalization.} $s_1=q(e_1;5^\circ)$ and $s_2=G(q(u_s;3\,\mathrm{px}),q(u_b;8\,\mathrm{px}))$.

\subsection{Phase Transitions and Melting}
\label{app:catalog_phase_transitions}

\paragraph{\normalfont\bfseries P24: Expansion when water freezes (Easy).}\mbox{}\par\nobreak
\noindent\textit{Definition.} Water freezes completely in a straight-sided vessel without material transfer or a change in cross-section.

\textit{Metrics.} M1 measures height expansion; M2 measures width stability:
\[
e_1=\frac{H_{\rm ice}}{H_{\rm water}}-\frac{1000}{917},
\qquad e_2=\frac{|W_{\rm final}-W_{\rm initial}|}{\max(W_{\rm initial},1\,\mathrm{px})}.
\]

\textit{Normalization.} $s_1=q(e_1;1.0)$ and $s_2=q(e_2;1.0)$.

\paragraph{\normalfont\bfseries P25: Floating freshwater ice melting in freshwater (Hard).}\mbox{}\par\nobreak
\noindent\textit{Definition.} Floating freshwater ice melts completely in a straight-sided freshwater vessel without material exchange; initial and final levels should agree.

\textit{Metrics.} M1 measures level change; M2 measures vessel-width change:
\[
e_1=\frac{y_{\rm before}-y_{\rm after}}{H_{\rm vessel}},
\qquad e_2=\frac{|W_{\rm after}-W_{\rm before}|}{W_{\rm before}}.
\]

\textit{Normalization.} $s_1=q(e_1;0.05)$ and $s_2=q(e_2;0.05)$.

\paragraph{\normalfont\bfseries P26: Melting floating ice containing a stone (Hard).}\mbox{}\par\nobreak
\noindent\textit{Definition.} Floating ice melts and releases an embedded dense stone, which sinks; the final water level should decrease.

\textit{Metrics.} M1 measures $e_1=(y_{\rm before}-y_{\rm after})/H_{\rm initial}$ after melting and a level change exceeding measurement uncertainty are observed. M2 is not defined.

\textit{Normalization.} $s_1=\mathbb{I}[e_1<0]$. Below-resolution changes remain unmeasured.

\paragraph{\normalfont\bfseries P27: Freshwater ice melting in saltwater (Hard).}\mbox{}\par\nobreak
\noindent\textit{Definition.} Freshwater ice melts and mixes with saltwater without material exchange; the final level should rise and no ice should remain.

\textit{Metrics.} M1 records $e_1=\Delta y/H_{\rm initial}$, where $\Delta y=y_{\rm before}-y_{\rm after}$. M2 measures $r_s=E_{\rm solid,terminal}/E_{\rm solid,initial}$, an image-based solid-evidence ratio.

\textit{Normalization.} $s_1=\mathbb{I}[\Delta y>\delta_y]$, where $\delta_y$ is level-change uncertainty in pixels. M2 is $s_2=\operatorname{clip}(1-r_s)$, overridden by $s_2=0$ if residual ice is confirmed in the final frame.

\paragraph{\normalfont\bfseries P28: Crushed ice versus one ice block (Hard).}\mbox{}\par\nobreak
\noindent\textit{Definition.} Equal masses of crushed ice and one intact ice block melt simultaneously in identical vessels.

\textit{Metrics.} M1 compares liquid heights measured from each vessel's bottom and divided by its height. A qualifying lead occurs when the crushed-ice side exceeds the intact-ice side by more than their summed localization uncertainties. M2 is not defined.

\textit{Normalization.} $s_1=1$ for a qualifying lead lasting at least $0.25\,\mathrm{s}$, and $s_1=0$ for at least $0.50\,\mathrm{s}$ of continuous jointly readable evidence without such a lead. Otherwise the metric remains unresolved; gaps are not interpolated.

\subsection{Electrostatics, Magnetism, and Electromagnetic Induction}
\label{app:catalog_electromagnetism}

\paragraph{\normalfont\bfseries P29: Eddy-current braking on an incline (Easy).}\mbox{}\par\nobreak
\noindent\textit{Definition.} Matched magnetic and nonmagnetic objects descend a copper incline; the magnetic object should arrive later and move more slowly.

\textit{Metrics.} M1 measures $r_t=(f_m-f_0+1)/(f_c-f_0+1)$, where $f_m,f_c$ are first-arrival frames at a common displacement and $f_0$ is the initialization frame. The common displacement is $80\%$ of the smaller maximum displacement over jointly valid observations. M2 measures speed ratio $r_v=v_m/v_c$.

\textit{Normalization.} $s_1=\mathbb{I}[r_t>1]$ and $s_2=\mathbb{I}[r_v<1]$.

\paragraph{\normalfont\bfseries P30: Magnet entry and indicator illumination (Easy).}\mbox{}\par\nobreak
\noindent\textit{Definition.} A magnet starts at rest, passes through a fixed coil connected to a bidirectional indicator, exits, and stops.

\textit{Metrics.} M1 combines binary evidence of magnet entry $E$ and confirmed indicator illumination during entry $L$, with a $0.5\,\mathrm{s}$ matching tolerance. An already-lit indicator qualifies if it remains lit during the matching interval. M2 is not defined.

\textit{Normalization.} $s_1=0.5\mathbb{I}[E]+0.5\mathbb{I}[L]$. Confirmed entry with an unreadable indicator receives the lower-bound score $0.5$, with uncertainty range $[0.5,1]$.

\paragraph{\normalfont\bfseries P31: Symmetric equilibrium of charged balls (Medium).}\mbox{}\par\nobreak
\noindent\textit{Definition.} Two identical charged balls on equal insulating strings repel and settle into a symmetric outward-separated configuration.

\textit{Metrics.} M1 measures string-angle asymmetry, $e_1=|\theta_l-\theta_r|$, in degrees. M2 measures displacement asymmetry, $e_2=|\ln(d_l/d_r)|$, using horizontal offsets from the apparatus centerline.

\textit{Normalization.} $s_1=q(e_1;90^\circ)$ and $s_2=q(e_2;1.0)$. An observed failure of outward separation, settling, or fixed equal-length strings sets both scores to zero.

\paragraph{\normalfont\bfseries P32: Final directions of two compass needles (Medium).}\mbox{}\par\nobreak
\noindent\textit{Definition.} A wire between two compasses is energized; their needles rotate and settle with opposing polar directions.

\textit{Metrics.} M1 measures median smallest angular separation $d$ in degrees over jointly readable observations among the final five frames, requiring at least two valid frames. M2 is not defined.

\textit{Normalization.} $s_1=(1-\cos(\pi d/180))/2$; opposite directions score one and aligned directions score zero.

\paragraph{\normalfont\bfseries P33: Closed versus split jumping rings (Medium).}\mbox{}\par\nobreak
\noindent\textit{Definition.} Matched apparatuses simultaneously excite closed and split aluminum rings, each free to move along its fixed core.

\textit{Metrics.} M1 measures $r_h=h_{\rm open}/h_{\rm closed}$, with each peak rise divided by that ring's initial outer diameter. A plateau or descent must establish each peak. M2 is not defined.

\textit{Normalization.} $s_1=\operatorname{clip}((1-r_h)/0.20)$. A reliably observed zero closed-ring rise scores zero, with the ratio undefined.

\paragraph{\normalfont\bfseries P34: Eddy-current damping of solid and slotted plates (Hard).}\mbox{}\par\nobreak
\noindent\textit{Definition.} Solid and slotted conducting plates with matched mass and inertia swing through equivalent magnetic fields from the same release angle.

\textit{Metrics.} M1 counts complete cycles $N_{\rm solid},N_{\rm slotted}$ in a common window of at least $2\,\mathrm{s}$, using cutoff $\max(1^\circ,0.1\operatorname{mean}(|\theta_{0,\rm solid}|,|\theta_{0,\rm slotted}|))$. Cycles are bounded by same-polarity events; truncated cycles are excluded. M2 is not defined.

\textit{Normalization.} $s_1=\mathbb{I}[N_{\rm solid}<N_{\rm slotted}]$. If $N_{\rm slotted}=0$, $s_1=0$ and the count ratio is undefined.

\subsection{Granular Media and Discharge Flow}
\label{app:catalog_granular}

\paragraph{\normalfont\bfseries P35: Scale consistency of sand-pile repose angles (Easy).}\mbox{}\par\nobreak
\noindent\textit{Definition.} Two unequal-size piles of the same dry sand receive thin streams at their peaks and maintain comparable repose angles.

\textit{Metrics.} M1 measures $e_1=|\alpha_{\rm small}/\alpha_{\rm large}-1|$. M2 measures left--right slope asymmetry in degrees:
\[
e_2=\max\!\left(|\alpha_{{\rm small},l}-\alpha_{{\rm small},r}|,
|\alpha_{{\rm large},l}-\alpha_{{\rm large},r}|\right).
\]

\textit{Normalization.} $s_1=q(e_1;1.0)$ and $s_2=q(e_2;90^\circ)$.

\paragraph{\normalfont\bfseries P36: Discharge of water and sand from funnels (Hard).}\mbox{}\par\nobreak
\noindent\textit{Definition.} Water and dry sand drain simultaneously from identical conical funnels with equal outlets and initial fill heights.

\textit{Metrics.} M1 fits $Q\propto h^\beta$ for each material using cone geometry and an integrated volume/head fit:
\[
e_1=|\beta_{\rm water}-0.5|+|\beta_{\rm sand}|.
\]
M2 is not defined.

\textit{Normalization.} $s_1=q(e_1;1.0)$.

\subsection{Surface Tension and Viscous Flow}
\label{app:catalog_interfaces}

\paragraph{\normalfont\bfseries P37: Capillary rise in tubes of different radii (Easy).}\mbox{}\par\nobreak
\noindent\textit{Definition.} Two initially dry glass tubes of different inner radii are lowered into one liquid reservoir; liquid should rise higher in the narrower tube.

\textit{Metrics.} M1 measures $d=\operatorname{median}(h_{\rm narrow}-h_{\rm wide})$ over paired terminal observations, where $h=y_{\rm reservoir}-y_{\rm meniscus}$. M2 is not defined.

\textit{Normalization.} $s_1=\mathbb{I}[d>\delta_{\rm resolution}]$, with $d$ and measurement resolution in pixels.

\paragraph{\normalfont\bfseries P38: Terminal settling of unequal spheres (Easy).}\mbox{}\par\nobreak
\noindent\textit{Definition.} Two spheres of the same material and different radii settle through glycerin, reaching terminal speed before bottom contact.

\textit{Metrics.} M1 measures
\[
e_1=\left|\frac{v_{\rm large}/v_{\rm small}}{(r_{\rm large}/r_{\rm small})^2}-1\right|.
\]
Each speed is measured over the longest qualified nonzero terminal-speed window, with earlier windows preferred in ties; the radius ratio must be at least $1.1$. M2 is not defined.

\textit{Normalization.} $s_1=q(e_1;1.0)$.

\paragraph{\normalfont\bfseries P39: Curvature of the partition between unequal soap bubbles (Medium).}\mbox{}\par\nobreak
\noindent\textit{Definition.} Two unequal near-spherical soap bubbles join and retain a visible separating film without rupture or detachment.

\textit{Metrics.} M1 independently fits the outer radii $r_s,r_l$ and signed partition radius $r_p$:
\[
E_t=\left|r_p\left(\frac{1}{r_s}-\frac{1}{r_l}\right)-1\right|.
\]
The residual $e_1$ is the frame-duration-weighted median of $E_t$ over uniquely identifiable partition intervals of at least three frames and $0.08\,\mathrm{s}$. M2 is not defined.

\textit{Normalization.} $s_1=q(e_1;1.0)$. A resolved nearly straight partition whose curvature uncertainty excludes the prediction is assigned the unbounded-error limit, $s_1=0$.

\paragraph{\normalfont\bfseries P40: Volume conservation in droplet coalescence (Medium).}\mbox{}\par\nobreak
\noindent\textit{Definition.} Two unequal free water droplets approach, touch, and merge into one near-spherical droplet.

\textit{Metrics.} M1 measures $e_1=|r_f^3/(r_1^3+r_2^3)-1|$ from fitted radii at a common image scale. M2 measures $e_2=|\mathcal{R}_f-(\mathcal{R}_1+\mathcal{R}_2)/2|$, where roundness is $\mathcal{R}=1-\operatorname{RMS}_{\rm radial}/r_{\rm fitted}$.

\textit{Normalization.} $s_1=q(e_1;0.10)$ and $s_2=q(e_2;0.10)$.

\section{Detailed Evaluation}
\label{app:detailed_evaluation}

Tables~\ref{tab:detailed_evaluation} and~\ref{tab:detailed_evaluation_part2} report the consistency score ($C$), physical score ($P$), and combined score ($S$) for all 40 tasks and eight video generation models.
The consistency score assesses temporal consistency and task observability, while the physical score summarizes task-specific measurements.
Physical evaluation is performed independently for all available videos, regardless of whether they pass the consistency gate.
The reported $P$ therefore includes videos with $C<80$ and is not set to zero solely because the gate rejects a video.
For each video, we compute
$S=0.15C+0.85P\,\mathbf{1}[C\ge80]$. The gate affects only the contribution of $P$ to $S$; the original $C$ and independently evaluated $P$ are retained.
All scores are expressed on a 0--100 scale and averaged over available seeds within each model--task pair, with the gate applied before averaging.
\begin{table*}[!htbp]
\centering
\newsavebox{\VDMAppendixBoxA}
\begin{lrbox}{\VDMAppendixBoxA}
\begin{minipage}{\textwidth}
\centering
\small
\setlength{\abovecaptionskip}{4pt}
\setlength{\belowcaptionskip}{5pt}
\caption{\textbf{Detailed task-level evaluation (part 1 of 2).}
Automatic consistency ($C$), independent physical ($P$), and composite ($S$) scores (0--100; $\uparrow$) for four models across 40 tasks.
The all-model mean is repeated in both tables.}
\label{tab:detailed_evaluation}
\begingroup
\definecolor{VDMCombined}{HTML}{F1F4F7}
\definecolor{VDMFamily}{HTML}{EFF1F4}
\definecolor{VDMEasy}{HTML}{EAF1E9}
\definecolor{VDMMedium}{HTML}{F5EEDC}
\definecolor{VDMHard}{HTML}{F5E6E6}
\definecolor{VDMSummary}{HTML}{E7EDF3}
\setlength{\tabcolsep}{3pt}
\renewcommand{\arraystretch}{1.10}
\fontsize{12}{14.5}\selectfont
\resizebox{\textwidth}{!}{%
\begin{tabular}{@{}l l c *{5}{r r >{\columncolor{VDMCombined}}r}@{}}
\toprule
\multicolumn{3}{l}{\textbf{Task}} & \multicolumn{3}{c}{\textbf{Seedance 2.5}} & \multicolumn{3}{c}{\textbf{MiniMax H3}} & \multicolumn{3}{c}{\textbf{Cosmos 3 Super}} & \multicolumn{3}{c}{\textbf{VBVR}} & \multicolumn{3}{c}{\textbf{All-model mean}} \\
\cmidrule(lr){4-6} \cmidrule(lr){7-9} \cmidrule(lr){10-12} \cmidrule(lr){13-15} \cmidrule(lr){16-18}
ID & Physical phenomenon & Level & $C$ & $P$ & $S$ & $C$ & $P$ & $S$ & $C$ & $P$ & $S$ & $C$ & $P$ & $S$ & $C$ & $P$ & $S$ \\
\midrule
\rowcolor{VDMFamily}\multicolumn{18}{l}{\textbf{1. Translational Motion and Collisions}} \\
P1 & Bounce-height decay & \cellcolor{VDMEasy}E & 100.00 & 89.13 & \textbf{90.76} & 100.00 & 86.10 & 88.19 & 100.00 & 49.96 & 57.46 & 0.00 & 10.00 & 0.00 & 78.12 & 58.14 & 57.19 \\
P2 & Free fall & \cellcolor{VDMMedium}M & 100.00 & 17.06 & 29.50 & 100.00 & 12.76 & 25.84 & 100.00 & 22.99 & \textbf{34.54} & 100.00 & 10.90 & 24.27 & 98.75 & 13.88 & 26.61 \\
P3 & Complementary-angle throws & \cellcolor{VDMMedium}M & 100.00 & 45.83 & 53.96 & 96.25 & 50.84 & \textbf{57.66} & 95.00 & 7.91 & 20.97 & 72.50 & 0.00 & 10.88 & 54.53 & 16.51 & 22.21 \\
P4 & Projectile motion & \cellcolor{VDMHard}H & 100.00 & 16.62 & 29.13 & 100.00 & 28.25 & \textbf{39.01} & 75.00 & 7.06 & 17.25 & 75.00 & 19.28 & 22.33 & 59.38 & 15.01 & 17.67 \\
P5 & Equal-mass collision & \cellcolor{VDMHard}H & 100.00 & 12.20 & 25.37 & 100.00 & 13.01 & \textbf{26.06} & 96.67 & 0.00 & 14.50 & 75.00 & 9.23 & 17.05 & 74.58 & 5.37 & 15.19 \\
\addlinespace[1.5pt]
\rowcolor{VDMFamily}\multicolumn{18}{l}{\textbf{2. Rolling, Friction, and Rigid-Body Statics}} \\
P6 & Mass-independent sliding & \cellcolor{VDMEasy}E & 100.00 & 60.00 & 66.00 & 100.00 & 97.52 & 97.89 & 100.00 & 98.18 & \textbf{98.46} & 100.00 & 42.50 & 51.12 & 93.75 & 60.35 & 62.71 \\
P7 & Hanging-chain equilibrium & \cellcolor{VDMEasy}E & 100.00 & 63.35 & 68.85 & 100.00 & 65.04 & 70.29 & 100.00 & 65.98 & 71.09 & 100.00 & 64.37 & 69.71 & 99.84 & 65.74 & 70.86 \\
P8 & Solid-sphere rolling & \cellcolor{VDMMedium}M & 100.00 & 71.25 & \textbf{75.56} & 100.00 & 36.26 & 45.82 & 97.50 & 16.67 & 28.79 & 98.75 & 19.67 & 31.53 & 95.16 & 25.53 & 35.97 \\
P9 & Solid sphere vs. hoop & \cellcolor{VDMMedium}M & 100.00 & 20.01 & 32.01 & 100.00 & 30.38 & 40.83 & 100.00 & 18.07 & 30.36 & 97.50 & 31.77 & \textbf{41.63} & 90.16 & 17.52 & 28.41 \\
P10 & Edge-pivot toppling & \cellcolor{VDMMedium}M & 100.00 & 22.67 & 34.27 & 100.00 & 60.62 & \textbf{66.53} & 100.00 & 22.39 & 34.03 & 25.00 & 1.18 & 4.76 & 75.00 & 18.56 & 27.02 \\
P11 & Rough-incline round trip & \cellcolor{VDMHard}H & 100.00 & 11.09 & 24.43 & 100.00 & 16.28 & \textbf{28.84} & 50.00 & 3.21 & 8.10 & 50.00 & 10.72 & 12.36 & 56.25 & 10.15 & 14.94 \\
\addlinespace[1.5pt]
\rowcolor{VDMFamily}\multicolumn{18}{l}{\textbf{3. Pendulum Motion and Oscillations}} \\
P12 & Pendulum period vs. mass & \cellcolor{VDMEasy}E & 100.00 & 76.96 & \textbf{80.41} & 100.00 & 67.74 & 72.58 & 100.00 & 62.54 & 68.16 & 75.00 & 57.17 & 56.28 & 93.75 & 46.52 & 53.09 \\
P13 & Large-angle pendulum & \cellcolor{VDMEasy}E & 100.00 & 50.30 & 57.76 & 100.00 & 76.33 & \textbf{79.88} & 100.00 & 56.25 & 62.81 & 75.00 & 51.44 & 43.59 & 93.44 & 55.48 & 58.33 \\
P14 & Pendulum period vs. length & \cellcolor{VDMEasy}E & 100.00 & 60.18 & 66.15 & 100.00 & 59.31 & 65.41 & 100.00 & 55.21 & 61.93 & 97.50 & 48.48 & 55.83 & 89.84 & 40.59 & 47.98 \\
P15 & Small-angle isochronism & \cellcolor{VDMMedium}M & 100.00 & 79.52 & \textbf{82.60} & 100.00 & 42.77 & 51.35 & 97.50 & 61.35 & 66.77 & 23.75 & 21.57 & 21.90 & 82.81 & 34.48 & 41.73 \\
\addlinespace[1.5pt]
\rowcolor{VDMFamily}\multicolumn{18}{l}{\textbf{4. Optics and Projective Geometry}} \\
P16 & Collinear-point cross-ratio & \cellcolor{VDMEasy}E & 100.00 & 77.76 & 81.10 & 100.00 & 79.96 & 82.97 & 100.00 & 79.28 & 82.39 & 100.00 & 81.09 & 83.92 & 84.06 & 68.79 & 66.77 \\
P17 & Light refraction & \cellcolor{VDMMedium}M & 100.00 & 59.51 & 65.59 & 100.00 & 59.62 & 65.68 & 100.00 & 48.68 & 56.38 & 100.00 & 60.22 & \textbf{66.18} & 99.69 & 34.82 & 44.55 \\
P18 & Light reflection & \cellcolor{VDMMedium}M & 100.00 & 95.47 & \textbf{96.15} & 75.00 & 70.07 & 49.82 & 0.00 & 29.99 & 0.00 & 97.50 & 95.44 & 95.75 & 43.12 & 41.39 & 33.73 \\
P19 & Projection concurrency & \cellcolor{VDMMedium}M & 100.00 & 61.34 & 67.14 & 72.50 & 40.87 & 45.62 & 100.00 & 39.63 & 48.69 & 97.50 & 64.72 & \textbf{69.64} & 76.88 & 28.44 & 35.71 \\
P20 & Refraction and total internal reflection & \cellcolor{VDMHard}H & 55.00 & 14.29 & 9.09 & 100.00 & 19.61 & 31.67 & 75.00 & 2.87 & 13.69 & 47.50 & 38.08 & 22.86 & 67.34 & 18.41 & 19.28 \\
\addlinespace[1.5pt]
\rowcolor{VDMFamily}\multicolumn{18}{l}{\textbf{5. Hydrostatics and Buoyancy}} \\
P21 & Communicating vessels & \cellcolor{VDMEasy}E & 100.00 & 96.48 & 97.01 & 100.00 & 97.84 & 98.16 & 100.00 & 97.29 & 97.70 & 100.00 & 98.83 & \textbf{99.01} & 100.00 & 95.86 & 96.48 \\
P22 & Floating-ice immersion & \cellcolor{VDMEasy}E & 100.00 & 60.66 & 66.56 & 100.00 & 80.26 & 83.22 & 100.00 & 75.08 & 78.82 & 100.00 & 82.70 & 85.29 & 98.75 & 72.19 & 76.18 \\
P23 & Liquid-surface orientation & \cellcolor{VDMMedium}M & 75.00 & 60.95 & 57.52 & 100.00 & 67.86 & \textbf{72.68} & 25.00 & 5.37 & 8.32 & 0.00 & 38.39 & 0.00 & 49.69 & 34.78 & 30.33 \\
\addlinespace[1.5pt]
\rowcolor{VDMFamily}\multicolumn{18}{l}{\textbf{6. Phase Transitions and Melting}} \\
P24 & Freezing-induced expansion & \cellcolor{VDMEasy}E & 100.00 & 93.30 & 94.30 & 100.00 & 95.55 & \textbf{96.21} & 75.00 & 88.11 & 66.69 & 22.50 & 93.06 & 24.38 & 52.81 & 89.05 & 48.97 \\
P25 & Ice melting: water level & \cellcolor{VDMHard}H & 100.00 & 94.84 & \textbf{95.61} & 0.00 & 0.00 & 0.00 & 100.00 & 0.00 & 15.00 & 0.00 & 0.00 & 0.00 & 28.12 & 11.86 & 14.30 \\
P26 & Ice with a stone: melting & \cellcolor{VDMHard}H & 100.00 & 0.00 & \textbf{15.00} & 0.00 & 0.00 & 0.00 & 25.00 & 0.00 & 3.75 & 0.00 & 0.00 & 0.00 & 22.97 & 0.00 & 3.45 \\
P27 & Freshwater ice in saltwater & \cellcolor{VDMHard}H & 100.00 & 37.50 & 46.87 & 0.00 & 0.00 & 0.00 & 100.00 & 50.00 & \textbf{57.50} & 0.00 & 0.00 & 0.00 & 28.12 & 12.50 & 13.52 \\
P28 & Crushed vs. intact ice & \cellcolor{VDMHard}H & 100.00 & 50.00 & \textbf{57.50} & 100.00 & 25.00 & 36.25 & 100.00 & 0.00 & 15.00 & 100.00 & 0.00 & 15.00 & 75.00 & 9.38 & 19.22 \\
\addlinespace[1.5pt]
\rowcolor{VDMFamily}\multicolumn{18}{l}{\textbf{7. Electrostatics, Magnetism, and Electromagnetic Induction}} \\
P29 & Eddy-current braking & \cellcolor{VDMEasy}E & 100.00 & 25.00 & 36.25 & 100.00 & 87.50 & \textbf{89.38} & 100.00 & 75.00 & 78.75 & 50.00 & 50.00 & 39.38 & 78.12 & 53.12 & 48.91 \\
P30 & Coil-induced light emission & \cellcolor{VDMEasy}E & 100.00 & 50.00 & 57.50 & 100.00 & 50.00 & 57.50 & 100.00 & 62.50 & \textbf{68.12} & 100.00 & 50.00 & 57.50 & 87.50 & 46.88 & 50.31 \\
P31 & Charged-sphere equilibrium & \cellcolor{VDMMedium}M & 100.00 & 98.05 & \textbf{98.34} & 100.00 & 96.75 & 97.24 & 100.00 & 0.00 & 15.00 & 100.00 & 0.00 & 15.00 & 96.41 & 24.35 & 35.16 \\
P32 & Final compass orientations & \cellcolor{VDMMedium}M & 100.00 & 25.21 & \textbf{36.43} & 100.00 & 0.06 & 15.05 & 100.00 & 0.00 & 15.00 & 100.00 & 3.54 & 18.01 & 100.00 & 7.08 & 21.02 \\
P33 & Closed vs. open jumping rings & \cellcolor{VDMMedium}M & 100.00 & 9.41 & 23.00 & 100.00 & 6.88 & 20.84 & 100.00 & 0.00 & 15.00 & 100.00 & 0.00 & 15.00 & 100.00 & 11.41 & 24.70 \\
P34 & Solid vs. slotted plate damping & \cellcolor{VDMHard}H & 100.00 & 25.00 & \textbf{36.25} & 100.00 & 0.00 & 15.00 & 100.00 & 0.00 & 15.00 & 0.00 & 0.00 & 0.00 & 77.50 & 3.12 & 14.28 \\
\addlinespace[1.5pt]
\rowcolor{VDMFamily}\multicolumn{18}{l}{\textbf{8. Granular Media and Discharge Flow}} \\
P35 & Sandpile angle scaling & \cellcolor{VDMEasy}E & 100.00 & 98.92 & \textbf{99.08} & 100.00 & 93.18 & 94.21 & 100.00 & 68.98 & 73.64 & 100.00 & 98.29 & 98.55 & 98.75 & 83.17 & 85.50 \\
P36 & Sand vs. water discharge & \cellcolor{VDMHard}H & 100.00 & 27.52 & \textbf{38.39} & 98.75 & 8.32 & 21.89 & 93.75 & 1.40 & 15.25 & 47.50 & 0.00 & 7.12 & 73.28 & 5.52 & 15.68 \\
\addlinespace[1.5pt]
\rowcolor{VDMFamily}\multicolumn{18}{l}{\textbf{9. Surface Tension and Viscous Flow}} \\
P37 & Capillary rise vs. diameter & \cellcolor{VDMEasy}E & 100.00 & 75.00 & 78.75 & 100.00 & 100.00 & \textbf{100.00} & 97.50 & 50.00 & 57.12 & 100.00 & 100.00 & \textbf{100.00} & 93.28 & 43.75 & 51.18 \\
P38 & Viscous settling speed & \cellcolor{VDMEasy}E & 100.00 & 61.04 & 66.88 & 100.00 & 61.47 & 67.25 & 100.00 & 62.94 & 68.50 & 100.00 & 64.52 & \textbf{69.85} & 93.44 & 59.11 & 62.45 \\
P39 & Bubble-film curvature & \cellcolor{VDMMedium}M & 100.00 & 0.00 & 15.00 & 100.00 & 47.15 & \textbf{55.08} & 100.00 & 12.12 & 25.30 & 100.00 & 16.22 & 28.79 & 100.00 & 20.07 & 32.06 \\
P40 & Droplet volume conservation & \cellcolor{VDMMedium}M & 100.00 & 50.94 & 58.30 & 100.00 & 33.56 & 43.52 & 100.00 & 20.54 & 32.46 & 50.00 & 67.62 & 37.07 & 77.50 & 32.06 & 34.29 \\
\midrule
\rowcolor{VDMSummary}\multicolumn{3}{l}{\textbf{Average}} & 98.25 & 51.11 & \textbf{57.76} & 91.06 & 49.12 & 54.89 & 90.07 & 35.44 & 42.46 & 69.44 & 37.52 & 37.79 & 78.44 & 34.77 & 38.95 \\
\bottomrule
\end{tabular}%
}
\endgroup
\end{minipage}
\end{lrbox}
\ifdim\dimexpr\ht\VDMAppendixBoxA+\dp\VDMAppendixBoxA\relax>.95\textheight
  \resizebox*{!}{.95\textheight}{\usebox{\VDMAppendixBoxA}}%
\else
  \usebox{\VDMAppendixBoxA}%
\fi
\end{table*}

\clearpage

\begin{table*}[!htbp]
\centering
\newsavebox{\VDMAppendixBoxB}
\begin{lrbox}{\VDMAppendixBoxB}
\begin{minipage}{\textwidth}
\centering
\small
\setlength{\abovecaptionskip}{4pt}
\setlength{\belowcaptionskip}{5pt}
\caption{\textbf{Detailed task-level evaluation (part 2 of 2).}
Automatic consistency ($C$), independent physical ($P$), and composite ($S$) scores (0--100; $\uparrow$) for four models across 40 tasks.
The all-model mean is repeated in both tables.}
\label{tab:detailed_evaluation_part2}
\begingroup
\definecolor{VDMCombined}{HTML}{F1F4F7}
\definecolor{VDMFamily}{HTML}{EFF1F4}
\definecolor{VDMEasy}{HTML}{EAF1E9}
\definecolor{VDMMedium}{HTML}{F5EEDC}
\definecolor{VDMHard}{HTML}{F5E6E6}
\definecolor{VDMSummary}{HTML}{E7EDF3}
\setlength{\tabcolsep}{3pt}
\renewcommand{\arraystretch}{1.10}
\fontsize{12}{14.5}\selectfont
\resizebox{\textwidth}{!}{%
\begin{tabular}{@{}l l c *{5}{r r >{\columncolor{VDMCombined}}r}@{}}
\toprule
\multicolumn{3}{l}{\textbf{Task}} & \multicolumn{3}{c}{\textbf{Wan 2.2}} & \multicolumn{3}{c}{\textbf{LingBot}} & \multicolumn{3}{c}{\textbf{Hunyuan 1.5}} & \multicolumn{3}{c}{\textbf{CogVideoX 1.5}} & \multicolumn{3}{c}{\textbf{All-model mean}} \\
\cmidrule(lr){4-6} \cmidrule(lr){7-9} \cmidrule(lr){10-12} \cmidrule(lr){13-15} \cmidrule(lr){16-18}
ID & Physical phenomenon & Level & $C$ & $P$ & $S$ & $C$ & $P$ & $S$ & $C$ & $P$ & $S$ & $C$ & $P$ & $S$ & $C$ & $P$ & $S$ \\
\midrule
\rowcolor{VDMFamily}\multicolumn{18}{l}{\textbf{1. Translational Motion and Collisions}} \\
P1 & Bounce-height decay & \cellcolor{VDMEasy}E & 100.00 & 83.13 & 85.66 & 100.00 & 39.62 & 48.67 & 100.00 & 80.04 & 83.03 & 25.00 & 27.14 & 3.75 & 78.12 & 58.14 & 57.19 \\
P2 & Free fall & \cellcolor{VDMMedium}M & 100.00 & 9.08 & 22.72 & 97.50 & 23.27 & 34.41 & 100.00 & 10.11 & 23.60 & 92.50 & 4.89 & 18.03 & 98.75 & 13.88 & 26.61 \\
P3 & Complementary-angle throws & \cellcolor{VDMMedium}M & 0.00 & 0.00 & 0.00 & 50.00 & 27.49 & 30.87 & 22.50 & 0.00 & 3.38 & 0.00 & 0.00 & 0.00 & 54.53 & 16.51 & 22.21 \\
P4 & Projectile motion & \cellcolor{VDMHard}H & 25.00 & 14.37 & 10.36 & 75.00 & 11.40 & 15.43 & 25.00 & 21.24 & 7.83 & 0.00 & 1.85 & 0.00 & 59.38 & 15.01 & 17.67 \\
P5 & Equal-mass collision & \cellcolor{VDMHard}H & 75.00 & 7.10 & 14.79 & 25.00 & 1.44 & 4.97 & 75.00 & 0.00 & 11.25 & 50.00 & 0.00 & 7.50 & 74.58 & 5.37 & 15.19 \\
\addlinespace[1.5pt]
\rowcolor{VDMFamily}\multicolumn{18}{l}{\textbf{2. Rolling, Friction, and Rigid-Body Statics}} \\
P6 & Mass-independent sliding & \cellcolor{VDMEasy}E & 100.00 & 42.33 & 50.98 & 100.00 & 50.00 & 57.50 & 100.00 & 24.78 & 36.06 & 50.00 & 67.50 & 43.62 & 93.75 & 60.35 & 62.71 \\
P7 & Hanging-chain equilibrium & \cellcolor{VDMEasy}E & 100.00 & 66.19 & 71.26 & 98.75 & 63.88 & 69.11 & 100.00 & 72.88 & \textbf{76.95} & 100.00 & 64.24 & 69.60 & 99.84 & 65.74 & 70.86 \\
P8 & Solid-sphere rolling & \cellcolor{VDMMedium}M & 100.00 & 23.99 & 35.39 & 100.00 & 4.44 & 18.77 & 97.50 & 26.45 & 37.11 & 67.50 & 5.47 & 14.78 & 95.16 & 25.53 & 35.97 \\
P9 & Solid sphere vs. hoop & \cellcolor{VDMMedium}M & 100.00 & 2.68 & 17.28 & 98.75 & 8.55 & 22.08 & 100.00 & 28.66 & 39.36 & 25.00 & 0.00 & 3.75 & 90.16 & 17.52 & 28.41 \\
P10 & Edge-pivot toppling & \cellcolor{VDMMedium}M & 100.00 & 33.21 & 43.23 & 75.00 & 3.68 & 14.38 & 100.00 & 4.70 & 18.99 & 0.00 & 0.00 & 0.00 & 75.00 & 18.56 & 27.02 \\
P11 & Rough-incline round trip & \cellcolor{VDMHard}H & 50.00 & 15.51 & 16.53 & 50.00 & 16.25 & 19.18 & 50.00 & 5.56 & 10.10 & 0.00 & 2.59 & 0.00 & 56.25 & 10.15 & 14.94 \\
\addlinespace[1.5pt]
\rowcolor{VDMFamily}\multicolumn{18}{l}{\textbf{3. Pendulum Motion and Oscillations}} \\
P12 & Pendulum period vs. mass & \cellcolor{VDMEasy}E & 100.00 & 24.19 & 35.56 & 75.00 & 10.07 & 19.25 & 100.00 & 43.63 & 52.08 & 100.00 & 29.86 & 40.38 & 93.75 & 46.52 & 53.09 \\
P13 & Large-angle pendulum & \cellcolor{VDMEasy}E & 100.00 & 45.61 & 53.77 & 100.00 & 37.55 & 46.92 & 100.00 & 68.84 & 73.52 & 72.50 & 57.52 & 48.41 & 93.44 & 55.48 & 58.33 \\
P14 & Pendulum period vs. length & \cellcolor{VDMEasy}E & 100.00 & 61.31 & \textbf{67.11} & 100.00 & 12.19 & 25.36 & 48.75 & 10.22 & 16.00 & 72.50 & 17.83 & 26.03 & 89.84 & 40.59 & 47.98 \\
P15 & Small-angle isochronism & \cellcolor{VDMMedium}M & 97.50 & 17.72 & 29.69 & 75.00 & 14.29 & 23.40 & 73.75 & 25.24 & 32.52 & 95.00 & 13.40 & 25.64 & 82.81 & 34.48 & 41.73 \\
\addlinespace[1.5pt]
\rowcolor{VDMFamily}\multicolumn{18}{l}{\textbf{4. Optics and Projective Geometry}} \\
P16 & Collinear-point cross-ratio & \cellcolor{VDMEasy}E & 100.00 & 62.50 & 68.12 & 47.50 & 83.47 & 43.56 & 100.00 & 86.28 & \textbf{88.34} & 25.00 & 0.00 & 3.75 & 84.06 & 68.79 & 66.77 \\
P17 & Light refraction & \cellcolor{VDMMedium}M & 100.00 & 50.52 & 57.94 & 97.50 & 0.00 & 14.62 & 100.00 & 0.00 & 15.00 & 100.00 & 0.00 & 15.00 & 99.69 & 34.82 & 44.55 \\
P18 & Light reflection & \cellcolor{VDMMedium}M & 0.00 & 0.00 & 0.00 & 22.50 & 0.00 & 3.38 & 50.00 & 40.19 & 24.73 & 0.00 & 0.00 & 0.00 & 43.12 & 41.39 & 33.73 \\
P19 & Projection concurrency & \cellcolor{VDMMedium}M & 97.50 & 20.97 & 32.45 & 77.50 & 0.00 & 11.62 & 47.50 & 0.00 & 7.12 & 22.50 & 0.00 & 3.38 & 76.88 & 28.44 & 35.71 \\
P20 & Refraction and total internal reflection & \cellcolor{VDMHard}H & 46.25 & 28.28 & 14.50 & 75.00 & 8.25 & 14.88 & 95.00 & 29.90 & \textbf{39.66} & 45.00 & 6.04 & 7.87 & 67.34 & 18.41 & 19.28 \\
\addlinespace[1.5pt]
\rowcolor{VDMFamily}\multicolumn{18}{l}{\textbf{5. Hydrostatics and Buoyancy}} \\
P21 & Communicating vessels & \cellcolor{VDMEasy}E & 100.00 & 91.88 & 93.09 & 100.00 & 96.94 & 97.40 & 100.00 & 94.63 & 95.44 & 100.00 & 93.00 & 94.05 & 100.00 & 95.86 & 96.48 \\
P22 & Floating-ice immersion & \cellcolor{VDMEasy}E & 100.00 & 83.37 & \textbf{85.86} & 97.50 & 63.92 & 68.96 & 100.00 & 61.56 & 67.32 & 92.50 & 70.00 & 73.38 & 98.75 & 72.19 & 76.18 \\
P23 & Liquid-surface orientation & \cellcolor{VDMMedium}M & 50.00 & 15.96 & 21.06 & 75.00 & 53.42 & 48.36 & 72.50 & 28.06 & 34.72 & 0.00 & 8.26 & 0.00 & 49.69 & 34.78 & 30.33 \\
\addlinespace[1.5pt]
\rowcolor{VDMFamily}\multicolumn{18}{l}{\textbf{6. Phase Transitions and Melting}} \\
P24 & Freezing-induced expansion & \cellcolor{VDMEasy}E & 0.00 & 95.76 & 0.00 & 100.00 & 84.04 & 86.43 & 25.00 & 90.48 & 23.71 & 0.00 & 72.09 & 0.00 & 52.81 & 89.05 & 48.97 \\
P25 & Ice melting: water level & \cellcolor{VDMHard}H & 0.00 & 0.00 & 0.00 & 25.00 & 0.00 & 3.75 & 0.00 & 0.00 & 0.00 & 0.00 & 0.00 & 0.00 & 28.12 & 11.86 & 14.30 \\
P26 & Ice with a stone: melting & \cellcolor{VDMHard}H & 0.00 & 0.00 & 0.00 & 58.75 & 0.00 & 8.81 & 0.00 & 0.00 & 0.00 & 0.00 & 0.00 & 0.00 & 22.97 & 0.00 & 3.45 \\
P27 & Freshwater ice in saltwater & \cellcolor{VDMHard}H & 0.00 & 0.00 & 0.00 & 0.00 & 0.00 & 0.00 & 0.00 & 0.00 & 0.00 & 25.00 & 12.50 & 3.75 & 28.12 & 12.50 & 13.52 \\
P28 & Crushed vs. intact ice & \cellcolor{VDMHard}H & 100.00 & 0.00 & 15.00 & 50.00 & 0.00 & 7.50 & 50.00 & 0.00 & 7.50 & 0.00 & 0.00 & 0.00 & 75.00 & 9.38 & 19.22 \\
\addlinespace[1.5pt]
\rowcolor{VDMFamily}\multicolumn{18}{l}{\textbf{7. Electrostatics, Magnetism, and Electromagnetic Induction}} \\
P29 & Eddy-current braking & \cellcolor{VDMEasy}E & 75.00 & 75.00 & 53.75 & 100.00 & 50.00 & 57.50 & 75.00 & 50.00 & 32.50 & 25.00 & 12.50 & 3.75 & 78.12 & 53.12 & 48.91 \\
P30 & Coil-induced light emission & \cellcolor{VDMEasy}E & 100.00 & 50.00 & 57.50 & 100.00 & 62.50 & \textbf{68.12} & 75.00 & 50.00 & 32.50 & 25.00 & 0.00 & 3.75 & 87.50 & 46.88 & 50.31 \\
P31 & Charged-sphere equilibrium & \cellcolor{VDMMedium}M & 100.00 & 0.00 & 15.00 & 72.50 & 0.00 & 10.88 & 98.75 & 0.00 & 14.81 & 100.00 & 0.00 & 15.00 & 96.41 & 24.35 & 35.16 \\
P32 & Final compass orientations & \cellcolor{VDMMedium}M & 100.00 & 25.08 & 36.32 & 100.00 & 1.66 & 16.41 & 100.00 & 0.00 & 15.00 & 100.00 & 1.11 & 15.95 & 100.00 & 7.08 & 21.02 \\
P33 & Closed vs. open jumping rings & \cellcolor{VDMMedium}M & 100.00 & 25.00 & 36.25 & 100.00 & 50.00 & \textbf{57.50} & 100.00 & 0.00 & 15.00 & 100.00 & 0.00 & 15.00 & 100.00 & 11.41 & 24.70 \\
P34 & Solid vs. slotted plate damping & \cellcolor{VDMHard}H & 100.00 & 0.00 & 15.00 & 50.00 & 0.00 & 7.50 & 100.00 & 0.00 & 15.00 & 70.00 & 0.00 & 10.50 & 77.50 & 3.12 & 14.28 \\
\addlinespace[1.5pt]
\rowcolor{VDMFamily}\multicolumn{18}{l}{\textbf{8. Granular Media and Discharge Flow}} \\
P35 & Sandpile angle scaling & \cellcolor{VDMEasy}E & 100.00 & 97.74 & 98.08 & 90.00 & 19.54 & 30.11 & 100.00 & 96.69 & 97.18 & 100.00 & 91.98 & 93.18 & 98.75 & 83.17 & 85.50 \\
P36 & Sand vs. water discharge & \cellcolor{VDMHard}H & 72.50 & 6.55 & 16.44 & 71.25 & 0.00 & 10.69 & 52.50 & 0.00 & 7.88 & 50.00 & 0.36 & 7.81 & 73.28 & 5.52 & 15.68 \\
\addlinespace[1.5pt]
\rowcolor{VDMFamily}\multicolumn{18}{l}{\textbf{9. Surface Tension and Viscous Flow}} \\
P37 & Capillary rise vs. diameter & \cellcolor{VDMEasy}E & 100.00 & 25.00 & 36.25 & 76.25 & 0.00 & 11.44 & 100.00 & 0.00 & 15.00 & 72.50 & 0.00 & 10.88 & 93.28 & 43.75 & 51.18 \\
P38 & Viscous settling speed & \cellcolor{VDMEasy}E & 75.00 & 63.58 & 50.86 & 100.00 & 59.89 & 65.91 & 75.00 & 47.23 & 51.39 & 97.50 & 52.17 & 58.97 & 93.44 & 59.11 & 62.45 \\
P39 & Bubble-film curvature & \cellcolor{VDMMedium}M & 100.00 & 32.07 & 42.26 & 100.00 & 20.48 & 32.41 & 100.00 & 32.49 & 42.62 & 100.00 & 0.00 & 15.00 & 100.00 & 20.07 & 32.06 \\
P40 & Droplet volume conservation & \cellcolor{VDMMedium}M & 72.50 & 28.56 & 26.31 & 100.00 & 55.30 & \textbf{62.00} & 97.50 & 0.00 & 14.62 & 0.00 & 0.00 & 0.00 & 77.50 & 32.06 & 34.29 \\
\midrule
\rowcolor{VDMSummary}\multicolumn{3}{l}{\textbf{Average}} & 75.91 & 33.11 & 35.66 & 77.66 & 25.84 & 32.25 & 75.16 & 28.25 & 31.97 & 50.00 & 17.81 & 18.81 & 78.44 & 34.77 & 38.95 \\
\bottomrule
\end{tabular}%
}
\endgroup
\end{minipage}
\end{lrbox}
\ifdim\dimexpr\ht\VDMAppendixBoxB+\dp\VDMAppendixBoxB\relax>.95\textheight
  \resizebox*{!}{.95\textheight}{\usebox{\VDMAppendixBoxB}}%
\else
  \usebox{\VDMAppendixBoxB}%
\fi
\end{table*}






\end{document}